\documentclass[]{svmult}

\usepackage{graphicx} 
\usepackage{makeidx}
\usepackage{multicol}

\usepackage{cite}
 
\usepackage{makeidx} 
\makeindex

\newfont{\myf}{pcrrc7t scaled 800}

\begin{document}

\title*{Problem Solving and Complex Systems} \author{Fr\'ed\'eric
  Guinand, Yoann Pign\'e\thanks{This work is partially
    supported by the French Ministry of Higher Education and
    Research.}}  \institute{Le Havre University (France)}
\maketitle

\begin{abstract}

  The observation and modeling of natural Complex Systems (CSs) like
  the human nervous system, the evolution or the weather, allows the
  definition of special abilities and models reusable to solve other
  problems.  For instance, Genetic Algorithms or Ant Colony
  Optimizations are inspired from natural CSs to solve optimization
  problems. This paper proposes the use of ant-based systems to solve
  various problems with a non assessing approach. This means that
  solutions to some problem are not evaluated. They appear as
  resultant structures from the activity of the system. Problems are
  mode\-led with graphs and such structures are observed directly on
  these graphs. Problems of Multiple Sequences Alignment and Natural
  Language Processing are addressed with this approach.
\end{abstract}
\vspace*{0.5in} \vspace*{-\baselineskip}

\section{Introduction}
\label{sec:intro}
The central topic of the work presented in that paper is to propose a
method for implicit building of solutions for problems modeled by
graphs.

Whatever the considered domain, physics, bio\-lo\-gy,
ma\-the\-ma\-tics, so\-cial scien\-ces, che\-mis\-try, com\-pu\-ter
scien\-ce...  there exist nu\-me\-rous exam\-ples of systems
exhi\-biting glo\-bal pro\-per\-ties that emerge from the
inter\-actions bet\-ween the en\-ti\-ties that com\-po\-se the system
itself.  Shoal of fishes, flocks of birds \cite{Chate_Gregoire_2004},
bacteria colonies \cite{Ben-Jacob_Levine_2004}, sand piles
\cite{Bak_1996}, cellular automata
\cite{Wolfram_1984,heudin98evolution}, protein interaction networks
\cite{Evry_2004}, city formation, human languages \cite{Walter_1994}
are some such examples.  These systems are called {\em complex
systems}\index{Complex Systems} \cite{Zwirn_2003}.  They are opened,
crossed by various flows and their compounds are in interaction with
themselves and/or with an environment that do not belong to the system
itself.  They exhibit a property of
self-organization\index{Self-Organization} that can be defined as an
holistic and dynamic process allowing such systems to adapt themselves
to the static characteristics as well as dynamic changes of the
environment in which their compounds move and act.
  
The work presented here aims at exploiting that self-organization
property for computing solutions for problems modeled by graphs.  The
central idea consists in the conception of an artificial complex
system whose entities move in and act on a graph (the environment) in
which we are looking for structures of special interest: solutions of
our original problem.  In order to answer our expectations, the
entities must leave some marks in the environment and these marks
should define expected structures.  In other words, we want to observe
a projection on the graph of the organization emerging at the level of
the complex system.  For that purpose, the considered complex system
has to be composed of entities able to move in the graph, able to
interact with each-other and with their environment. The effects of
this last kind of interactions materialize the projection of the
organization of entities onto the environment.

In ethology,\index{Ethology} structures and organizations produced by
animal societies are studied according to two complementary points of
view \cite{Theraulaz_1997}.  The first tries to answer to the question
{\em why} while the second focuses on the question {\em how}. For the
latter, the goal is to discover the link that exists between
individual behavior and the structures and collective decisions that
emerge from the group.  In this context, self-organization is a key
concept. It is indeed considered that the global behavior is an
emergent process that results from the numerous interactions between
individuals and the environment.  More precisely, when animal
societies are considered, three families of collective phenomena may
be distinguished: (a) spatio-temporal organization of individual
activities, (b) individual differentiation and (c) phenomena leading
to the collective structuration of the environment. This latter
phenomenon mainly occur with wasp, termite and ant societies.  This
explains probably why for about one decade, ants have inspired so much
work on optimization \cite{Bonabeau_Dorigo_Theraulaz_1999}.

The approach described in this paper is closely related with the
question {\em how}.  The more promising collective phenomenon for
reaching our aim is the third one: collective structuration of the
environment.  Knowing previous works on ant colonies and more
precisely on Ant System,\index{Ant-based Systems} artificial ants
appear obviously as excellent candidates for representing the basic
entities of our artificial complex system.

Ants are mobile entities, they can interact directly using antennation
and/or trophallaxy, or non directly, using the so-called
stigmergy\index{Stigmergy} mechanism \cite{Grasse_59}.  Stigmergy may
be sign-based or sematectonic.  Sign-based stigmergy is illustrated by
ants dropping phe\-ro\-mones along a way from the nest to a food
source.  Sematectonic stigmergy is based on a structural modification
of the environment, has illustrated by termites building their nest
\cite{Grasse_1984}.  In its simplest version, an ant algorithm follows
three simple rules.  In order to simplify, let us consider that the
algorithm operates on a graph: (1) each ant drops a small quantity of
phe\-ro\-mone along the path it uses; (2) at a crossroads, the choice
of the ant is partially determined by the quantity of phe\-ro\-mones
on each outgoing edge: the choice is probabilistic and the larger the
quantity of phe\-ro\-mones on one edge, the larger the probability to
choose this edge; (3) phe\-ro\-mones evaporate with the time.  During
the process, some edges are more frequently visited than other ones,
making appear at the graph level some paths, groups of edges/vertices,
or other structures resulting from the local and indirect interactions
of ants.

Artificial ant colonies have been widely studied and applied to
optimization problems\index{Optimization}.  The way they have been
used makes ant colonies belonging to the general class of
metaheuristics\index{Metaheuristic}.  This term refers to high level
strategies that drive and modify other heuristics in order to produce
solutions better than what could be expected by classical approaches.
A metaheuristic may be defined as a set of algorithmic concepts that
can be used for the definition of heuristic methods applicable to a
large spectrum of problems \cite{ACO_2004}.  A heuristic rests on a
scheme that may be a constructive method or a local search method.  In
the former case, one solution is built using a greedy strategy, while
a local search strategy iteratively explores the neighborhood of the
current solution, trying to improve it by local changes.  These
changes are defined by a neighborhood structure that depends on the
addressed problem.  In \cite{ACO_2004}, the authors note that
artificial ants may be considered as simple agents and that the good
solutions correspond to an emergent property resulting from the
interactions between these cooperative agents.  In addition, they
describe an artificial ant in ACO (Ant Colony Optimization)

as a stochastic building process that builds step by step a solution
by adding opportunely elements to the partial solution.  The
considered problems are expressed in a classical way, with the
definition of a space of solutions, a cost function that should be
minimized and a set of constraints.  Then, each ant possesses an
evaluation function and this evaluation drives the process.

Particle Swarm Optimization (PSO) is another kind of optimization
method based on the collective behavior of simple agents. It is close
to ant colonies, has been proposed and described in
\cite{Kennedy_2001}.  But once again, in PSO, each particle owns an
evaluation function and knows what is the space of solutions.  Each
particle is characterized, at time $t$ by its position, its speed, a
neighborhood, and its best position according to the evaluation
function since the beginning of the process.  At each step, each
particle moves and its speed is updated according to its best
position, the position of the particle belonging to the neighborhood
showing the better result to the evaluation function.

Our approach, while belonging to the class of po\-pu\-lation-based
methods\index{Population-based Methods}, is different than both ACO
and PSO.  In our mind, depending on the model of the problem and on
the characteristics of the ants, the role of each individual ant is
not to compute a solution (neither complete, nor partial), but the
solution has to be observable into the environment as the result of
the artificial ants actions.  That may be considered as a kind of
implicit optimization.  We argue that it is possible to compute
non-trivial solutions to problems without using any global evaluation
function.  In what we have done, neither the ants, nor the elements of
the environment in which they move evaluate the result of their
interactions during the process.  The motivations for such an approach
are multiple.  Building a global evaluation function
\index{Evaluation Function}\index{Unpredictability}
 is sometimes not possible or would required an effort as
important as solving the problem itself.  Sometimes, the evaluation of
the function is not conceivable during the resolution process because
of the dynamics of the problem.  Some other problems cannot be
easily expressed as optimization problems, as in the example
developed in Sect.  \ref{sec:nlp}.
 
That's the reason why the behavior of our artificial ants remains
simple and only depends on local information.  Depending on the
application, we can have one or several collaborating or competing
colonies.  During the process, each ant can drop phe\-ro\-mones,
interact with other ants, and modify its environment.  The choice
between several edges for proceeding with the path depends on the
quantity of phe\-ro\-mones dropped by ants belonging to the same
colony (attraction) and on the quantity of phe\-ro\-mones dropped by
ants of another colony (repulsion).  The modification of the
environment may be the creation of a vertex or the deletion of an edge
on the graph.\\

In the sequel, the general approach is more precisely stated in
Sect. \ref{sec:approach}.  It is illustrated by two case studies: the
bioinformatic problem of multiple sequence alignment
(Sect. \ref{sec:msa}) and the search for interpretation trails in
texts written in natural languages (Sect. \ref{sec:nlp}).
Sect. \ref{sec:analysis} exposes some difficulties facing this
approach, particularly the identification of relevant parameters and
characteristics and the sensitivity of the method to critical
parameters. To finish, a conclusion draws some short and mid-term
perspectives.

\section{General Approach}
\label{sec:approach}

\begin{figure}[htb]
\centering
  \includegraphics[width=0.7\textwidth]{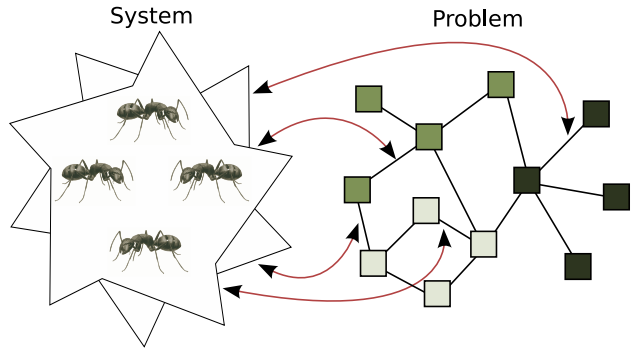}
  \caption{The general model: The system evolves on the graph of the
    problem, modifies it and is influenced by its own modifications.
    Finally solutions are directly observed on the graph.}
  \label{fig:userSystemProblem}
  \end{figure}

As a starting point, we consider problems that may be modeled by a
graph\index{Graphs}.  Evolutionary Computation excepted, in classical
metaheuristics (ACO, PSO, GAs, tabu search based methods, GRASP),
solutions are explicitly computed and one unavoidable step is the
evaluation of these solutions in order to drive further
investigations.  In EC the individuals are not necessarily solutions
of the considered problem, but it is always possible to build a
solution from the set of individuals.  This feature is not required
for us.  Indeed, our approach relies on the report that in many cases
solutions correspond to structures in the graph.  Such structures can
be paths, set of vertices, set of edges, partitions of the graph... or
any other group of graph elements.  We consider this graph as the
environment, into which many moving and active entities (our
artificial ants) are born, operate, and (sometimes) die.  They have
the knowledge of neither the environment, nor the final goal of the
system, nor the objective function.  Their actions are not driven by
the problem, only their characteristics may depend on it.  Each ant
can perform only two actions: to move and to modify the environment.
Moving depends on local characteristics of the environment and
sometimes on an additional basic goal (turning back nest for
instance).  Modifications of the environment is used for both purpose:
indirect communications between ants and structures forming in the
environment.  Modifications may be of two types, corresponding to the
two stigmergy types: sign-based (the phe\-ro\-mones) and sematectonic
(changes affect graph structure).  In addition, the modifications of
the environment have a feedback effect on the behavior of artificial
ants as illustrated on Fig. \ref{fig:userSystemProblem}.

As mentioned in the introduction, artificial ants societies are
complex systems, indeed, they are able to exhibit a self-organizing
property that may be translated into, for instance, a set of edges
with a large volume of phe\-ro\-mones, a set of vertices of the same
color... as many structures that could be solutions of some problems
in relation with the original graph.

The general process for problem-solving using a complex system as
described in the sequel is made of three steps: (1) identification of
the structures in the graph that can be associated to solutions of the
original problem, (2) choice of the features required by the system to
achieve the goal (that is building of structures), (3) critical
parameters tuning. Let us now enter into details.

\subsection{Structures Identification}
\index{Graphs!Structures|see{Emergence}} 
\index{Emergence}

The first step, before the construction of the system itself is to
identify, for the graph modeling the considered problem, the
structures that are solutions.

Let us illustrate this with some examples:\\
\begin{table} \centering
\caption{}
\begin{tabular}{l l} \hline\noalign{\smallskip} {\bf Problem description} & {\bf Relevant structure} \\ 
\hline Mapping & sets of vertices \\ 
Routing & set of paths \\ 
Partitioning & sets of vertices\\ 
TSP & circuits \\ 
Shortest path & paths \\ 
Multiple sequence alignment & sets of edges \\ 
DNA-Sequencing & paths \\
\hline\noalign{\smallskip}
\end{tabular}
\end{table}

The problem of mapping consists in allocating sets of tasks to sets of
resources. At the end of the process, each resource is allocated a set
of tasks. With respect to the graph this corresponds to sets of
vertices gathered according to their allocated resources.  Multiple
sequence alignment will be extensively described in Sect.
\ref{sec:msa}, and Word Sense Disambiguation in Sect. \ref{sec:nlp}.
For DNA-Sequencing, there exists several graph formulations. In all
cases, a solution corresponds to a path in the graph. According to the
model, the path may be either a Hamiltonian path, an Eulerian path or
a constrained path.  In the case of multi-objective problems, the
solutions can be presented as a set of paths or subgraphs that in the
best case all belong to the Pareto front.

However, a given problem may be modeled using different graph
formulation, thus, it is unlikely to find only one association between
a problem and a graph structure.

\begin{figure}[htb]
  \centering
  \includegraphics[width=0.4\textwidth]{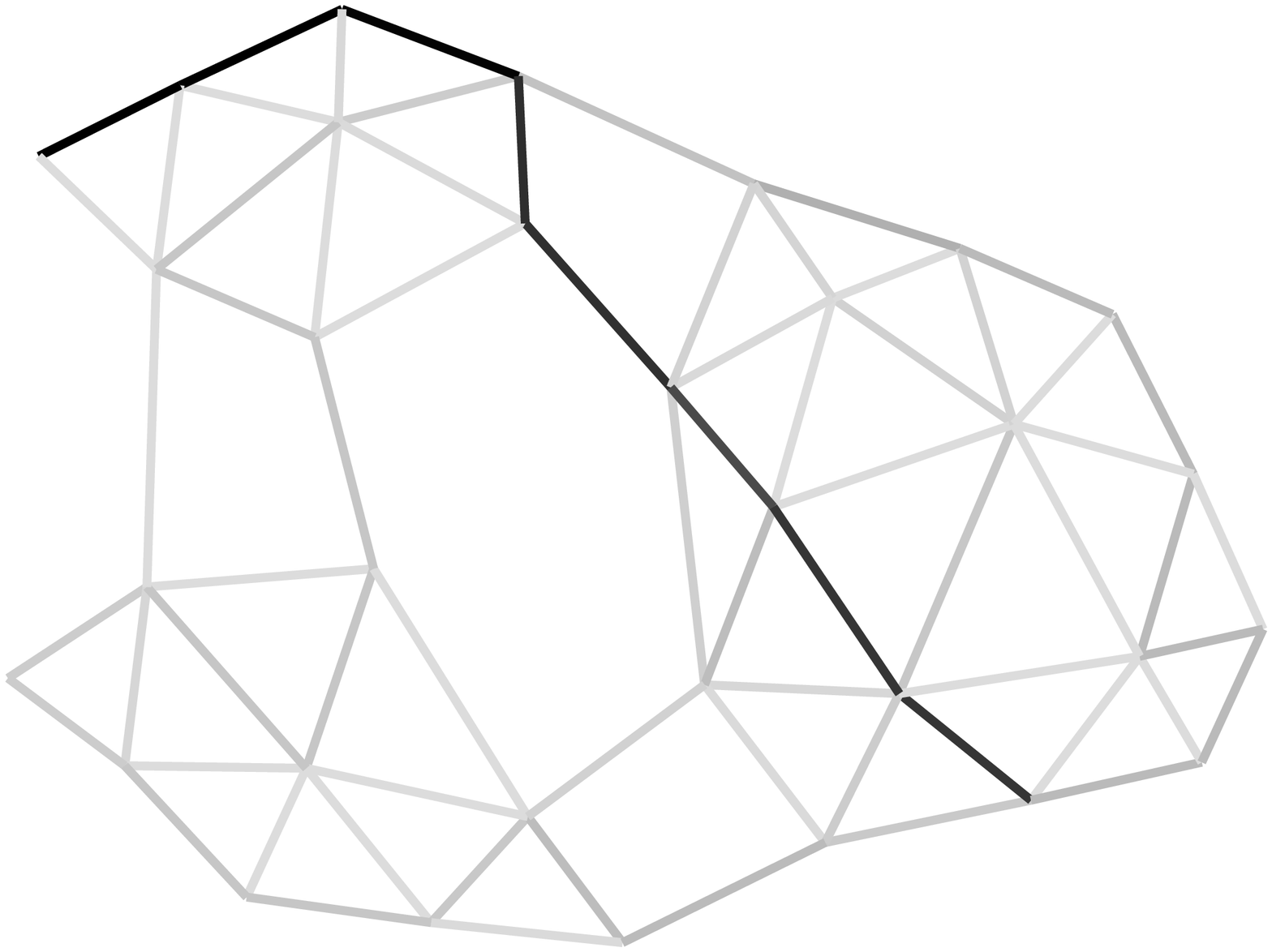} 
\hfill
\includegraphics[width=0.5\textwidth]{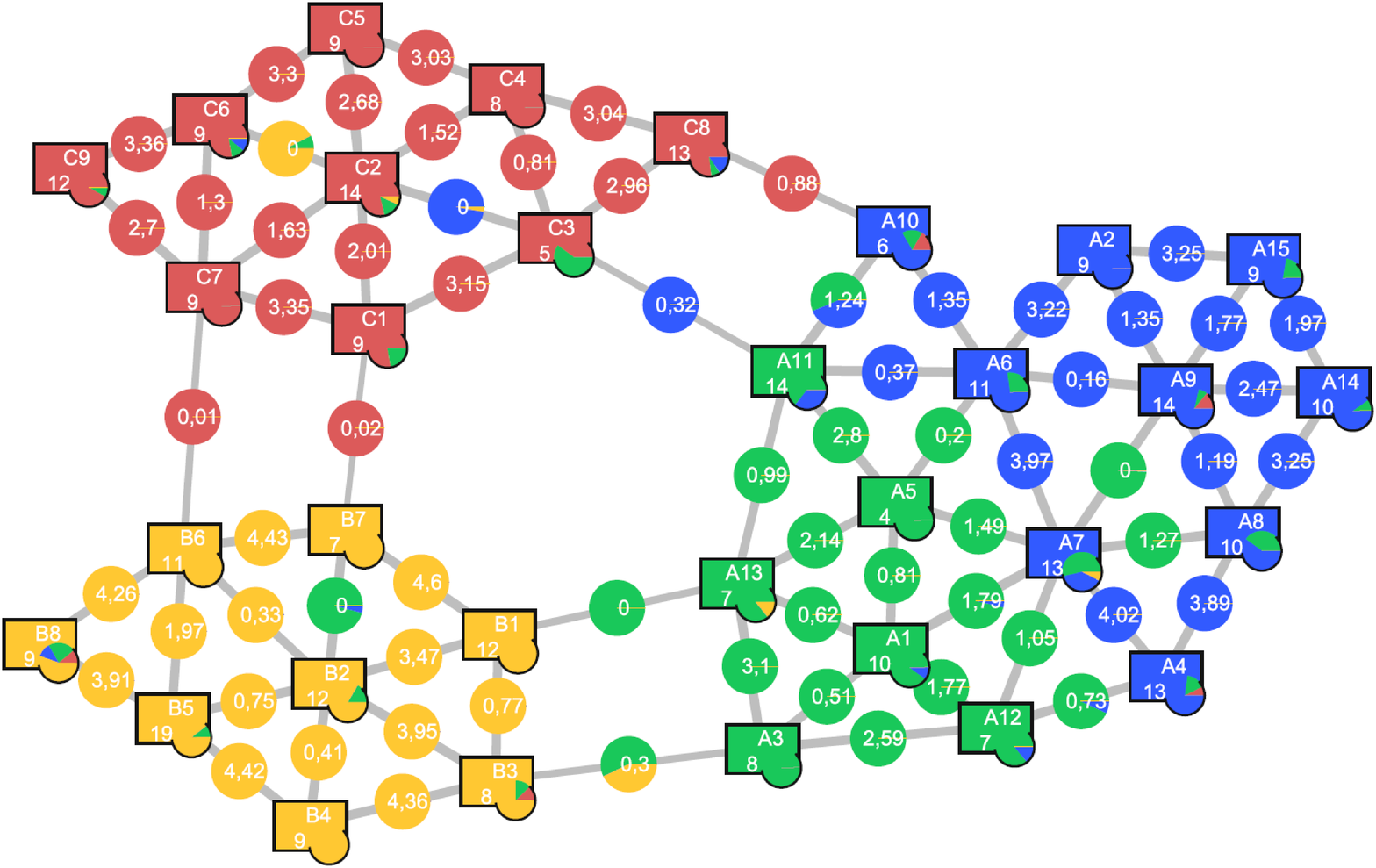}
\caption{Example of structures solutions of two different problems.}
\label{fig:structures}
\end{figure}

Figure \ref{fig:structures} illustrates how such structures may be
observable directly by a user. The first picture (left-hand side) is a
path built by ants for the shortest path problem. The original problem
associated to the second picture is dynamic load balancing. This time,
the structures are sets of colored vertices, to each color corresponds
a resource.  The method used for obtaining the right-hand side graph
is described in the same volume \cite{ANTS_2004}

\subsection{The Model} 
\index{Ant-based Systems|(}

Our choice for a complex system based on artificial
ants was their ability
to modify their environment.  Moreover, ant-based systems have an
additional property since the death of any ant do not impede the
system to work.  Finally, a lot of ants may be used together since
they don't directly communicate, but they use the environment instead,
thus, the overhead entailed remains acceptable, and the whole system
is easily distributed on a computer system (a grid for example) as
soon as the environment (the graph) is itself distributed.\\
 
As it is classically defined complex
systems (CSs) are led by feedback loops\index{Feedback Loops}.
Positive feedback amplify the evolution while negative feedback reduce
it.  Ant-based systems are bound to the same mechanisms.  Positive
feedback takes place with stigmergy.\index{Stigmergy}  
When performing walks in the
graph, ants modify their environment by structural changes in the
graph (sematectonic stigmergy) or by laying phe\-ro\-mones down the
graph (sign-based stigmergy).  This modification are information for
other ants that will later visit the same parts of the graph.  These
ants will be attracted by the environment's information and will
deposit their own information, attracting more and more ants into a
positive feedback loop.  If nothing stops the mechanism, the system
freezes and cannot evolve anymore. To prevent the system from
stagnating, negative feedback loops are used. In our approach, it is
obtained with evaporation or erosion processes.  In the nature, real
phe\-ro\-mone do evaporate.  This is a brake to previous stygmergic
mechanisms.

The ant-based complex system has three kinds of elements: the nests,
the colonies and the ants.  The conception of the system takes place
at two levels: the colony level and the ant level, the nests are only
positions from which the ants belonging to the corresponding colony
begin their walks.

\subsubsection{Nests and Colonies}


The system may be constituted by one or several colonies.  The choice
of the number of colonies is mainly problem-driven.  Indeed, if we
need only one structure, in most cases one colony is enough. But, in
case of solutions made of several structures determined as the result
of a competition (partitioning or mapping for instance), then, several
colonies should be necessary.  However, sometimes, it is necessary
that ants are born everywhere in the graph, in such particular cases,
there is a one-to-one mapping between nests and vertices. \\

One or more colonies of ants may operate into the environment. The
nest of the colony can take place in the graph on one particular
vertex, if the original problem defines a starting point.  This is the
case for instance when the structure is a path defined by a fixed
starting point. However, there are also scenarios for which all
vertices are nests, each corresponding to a distinct colony. In this
case ants are competing for accessing some resources.  It is also
possible to design systems in which each vertex is a nest belonging to
a supercolony (in the sense of unicoloniality \cite{Giraud_et_al_2002}
(Argentine ants in southern Europe)).\\

In all cases, an ant stemmed from one given colony inherit from this
colony a set of characteristics (eventually empty), a color, a
sensitivity to phe\-ro\-mones... More generally, colonies
differentiation allow the use of different heuristics on the same
problem.  A wide range of global system behavior rise from this.  One
of the most interesting properties is the use of ant differentiation
so as to raise competition or collaborative mechanisms.  These
mechanisms are useful when dealing with multiple objective problems or
when the construction of clusters or subgraphs are desired.

\index{Colored Ants}
To obtain competition, it is common to assign a colony a color. Ants from
one particular color lay down the environment "colored" phe\-ro\-mones
of the same color. Ants are attracted by the phe\-ro\-mones with their
color and are repulsed by other colored phe\-ro\-mones.  this general
behavior lead to the coloration of disconnected parts of the graph
like clusters or subgraphs.

To obtain collaboration, the same mechanism is used. Ants deposit
phe\-ro\-mones of their color and are attracted by it but there is no
repulsion with other colors. This mechanism is ideal when dealing with
multiple objective problems. One objective represent one color. Each
colony enhances its objective while being attracted by all the
phe\-ro\-mones.

\subsubsection{Ants} 
 
First, general characteristics of ants are studied.  Then details
about the way pheromone trails are constructed and how ants move are
considered.

\paragraph{General Characteristics}
 
Each ant is autonomous, and independent from the others, it can thus
be executed concurrently and asynchro\-nously.  Each ant is reactive
to the environment and its local neighborhood is defined as the set of
adjacent vertices and edges of the vertex the ant is located on. Local
information available from this set of vertices and edges constitute
the basic material for the ant to prepare its next move.  These
information are of two kinds: information belonging to the problem
itself if the problem needs a valuation of vertices and edges.  The
topology of the graph can also indicates constraints from the problem
(degree of vertices, directed versus non-directed graph).  Second, the
information raised from the activity of the ant colony: phe\-ro\-mones
and other data deposit on the edges and on the vertices.

Depending on the considered problem, ants are assigned
characteristics. These characteristics are more or less problem
specific. Some of them can be used with multiple problems and deserve
a presentation:
\begin{itemize}
\item The assignment of a color to the ants is useful when dealing
with partitioning problems or multiple objective problems.
\item A tabu list prevent ants from visiting twice an already visited
vertex.  The size of this list is relevant for many problems.
\item A sensitivity to phe\-ro\-mones parameter is usually used to
find a balance between the reinforcement of phe\-ro\-mone trails and
the diversification of the search. When ants are too much attracted by
phe\-ro\-mones they follow already defined trails and don't try to
visit unexplored parts of the graph. On the contrary, if they don't
pay enough attention to the phe\-ro\-mones, no stigmergy is available.
\end{itemize}

\paragraph{Moves and Pheromones} 
 
Every single ant's general aim is to move in the graph and to lay down
some phe\-ro\-mones during their moving.  These walks are led by local
information and constraints found in the environment.

Ants lay down phe\-ro\-mone trails on their way. The quantity of
phe\-ro\-mone deposited is a constant value or calculated with a local
rule but is never proportional to any evaluation of a constructed
solution. Pheromone is persistent: it continues to exist after the
ant's visit.  This phe\-ro\-mone will influence other ants in the
future, indeed, among the set of adjacent edges of one particular
vertex $v$, the larger the quantity of phe\-ro\-mone on the adjacent
edge the more attractive this edge will be for ants located on $v$.

The choice of the next vertex to move to is done locally for each
vertex as follows: Let consider a vertex $i$ who's neighbors vertices
are given by the function $neighbors(i)$ as a set of vertices.  The
probability $P_{i,j}$for one ant located on vertex $i$ to move to
neighbor vertex $j$ is:
\begin{equation} P_{i,j}=\frac{\tau_i}{\sum_{k \in
neighbors(i)}\tau_k}
	\label{eq:transRule}
\end{equation}

Different methods can be used to perform the phe\-ro\-mone lay
down. The nature of the lay down and the nature of the phe\-ro\-mones
themselves depends on the problem.

Depending on the kind of solution that is expected, these moves will
raise different structures.
 
\begin{itemize}
\item If the expected structure is a path or a set of paths in the
graph defined by a source vertex and a destination vertex, then ants
moves define paths.  The Pheromone deposit can be done when going to
the destination and when returning to the nest or when returning only.

 \item If the structures looks like clusters or subgraphs, then the
moves don't define paths. Ants have no destination but their behavior
is not necessarily different from the previous one. But the existence
of concurrent colonies may restrict the movements of ants and may
entail a kind of gathering of ants of the same color in the same
region.  The phe\-ro\-mone deposit will be performed at each step. If
ants are assigned a color, then the phe\-ro\-mones they lay down will
be colored with the same color.
\end{itemize}

\subsubsection{General Consideration about the Model}

The motivation of developing such a system comes from its three main
properties: self-organization, robustness and flexibility.
\index{Self-Organization}
\index{Robustness}
\index{Flexibility} 
The system do
not explicitly build structures, instead of that, structures appear as
the results of the interactions between ants and between ants and the
environment. This is a consequence of the self-organization property.

The system do not compute explicitly solutions but rather relies on
the notion of structures, thus, when changes occur within the
environment, the whole process remains the same since no objective
function is evaluated by ants, in other words, the behavior of ants
changes in no way. Thus, intrinsically, this approach owns a
robustness property.

Finally, the system is flexible. Indeed, ants are autonomous and
independent entities. If one or several ants disappear from the
system, this one keeps on working. This property is very interesting
because it allows a way of working based on flows of entities stemmed
from colonies that may be subject to mutations (practically changes in
some characteristics).

\index{Ant-based Systems|)}

\vfill
\pagebreak

\section{Multiple Sequence Alignment}
\label{sec:msa}

Multiple Sequence Alignment (MSA) 
\index{Multiple Sequence Alignment|(}  
is a wide-ranging problem. A lot of
work is done around it and many different methods are proposed to deal
with \cite{taylor88,thompson94,barton87}.  A subset of this problem,
called multiple block alignment aims at aligning together highly
conserved parts of the sequences before aligning the rest of the
sequences. Here, the problem isn't considered as a whole. The focus is
done on the underlining problem of MSA that is the alignment of blocks
between the sequences.

\subsection{Description of the Problem}
\label{sec:msa_pb}

Multiple alignment is an inescapable bioinformatic tool, essential to
the daily work of the researchers in molecular biology. The simple
versions come down to optimization problems which are for the majority
NP-hard. However, multiple alignment is very strongly related to the
question of the evolution of the organisms from which the sequences
result. It is indeed allowed that the probability of having close
sequences for two given organisms is all the more important as these
organisms are phylogenetically close. One of the major difficulties of
multiple alignment is to determine an alignment which, without
considering explicitly evolutionary aspects, is biologically relevant.

Among the many ways followed to determine satisfactory alignments, one
of them rests on the notion of block.  A factor is a substring present
in several sequences and should correspond to strongly preserved zones
from an evolutionary point of view. This particular point makes this
approach naturally relevant from the biological point of view.
Building blocks consists in choosing and gathering identical factors
common to several sequences in the more appropriate way.  A block
contains at most one factor per sequence and one given factor can only
belong to one block. The construction of blocks is one step of the
full process of multiple sequence alignment and the choice of factors
for building blocks is the problem we address.

\subsection{Proposed Solution}
\label{sec:msa_proposedSolution}

The general scheme for this kind of the method follows three steps:
\begin{enumerate}
\item Detection of common subsets (factors) in the set of sequences.
\item Determination of groups of common factors (blocks) without
  conflict between them.
\item Alignment of the rest of the sequences (parts of sequences that
  don't belong to the set of blocks).
\end{enumerate}

The work done by the ant colony deals with the second step. The
problem must be modeled as a graph. It will constitute the environment
that the ants will be able to traverse. Solutions are observed in the
graph as a set of relevant edges that link factors into blocks.

\subsection{Graph Model}
\label{sec:msa_graphModel}

\begin{figure}[htb]
    \centering
    \includegraphics[width=0.7\textwidth]{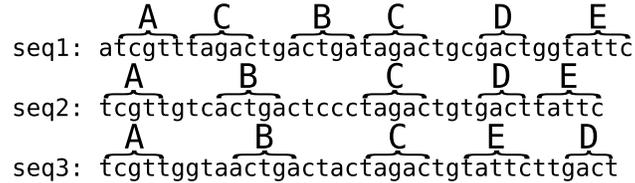}
    \caption{Sequences conversion}{This is an alignment composed of
      three sequences.  Common subsequences of these sequences which
      are repeated are labeled. After the conversion, the alignment is
      described as sequence~1~=~ACBCDE, sequence~2~=~ABCDE and
      sequence~3~=~ABCED.}
  \label{fig:sequences}
\end{figure}

As only factors of sequences are manipulated, each DNA or protein
sequence considered is reduced to the list of its common factors with
the other sequences of alignment. Figure \ref{fig:sequences}
illustrates such a conversion.

Commonly, a graph is used to figure out the environment in which ants
are dropped off. A multiple alignment is often represented by laying
out the sequences one under the other and by connecting the identical
factors together. Factors and their relations in the alignment can be
represented as graphs. The factors are the vertices and the edges
between these factors are the edges of the graph.

A group of identical factors (with the same pattern) form a {\bf
  factor graph}.  The quantity of factor graphs is equal to the number
of different factors.  That is to say, given a factor 'A', the factor
graph 'A' is a complete graph in which all the edges linking the
factors belonging to the same sequence are removed.

A group of blocks is said to be {\bf compatible} if all of these
blocks can be present in the alignment with no cross and no overlap
between them.

If two blocks cross each other, one of them will have to be excluded
from the final alignment in order to respect the compatibility of the
unit. In a graph representation, one can say that the crossings
between the edges are to be avoided.

\begin{figure}[htb]
  \begin{minipage}[b]{1.0\linewidth}
    \centering
    \includegraphics[width=0.7\textwidth]{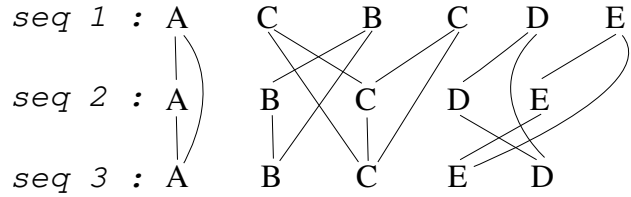}
  \end{minipage}
  \caption{The same alignment as in Fig. \ref{fig:sequences}. Here,
    the factors and the edges between them are presented.}
  \label{fig:sequences2}
\end{figure}

This representation makes it possible to locate the conflicts between
the factors which will prevent later the formation of blocks. Indeed,
as in Fig. \ref{fig:sequences2}, the factors 'D' and 'E' of sequence 2
are linked to the factors 'D' and 'E' of sequence 3. These edges cross
each other, which translates a conflict between 2 potential blocks
containing these factors.

\begin{figure}[htb]
  \begin{minipage}[b]{\linewidth}
    \centering
    \includegraphics[width=0.7\textwidth]{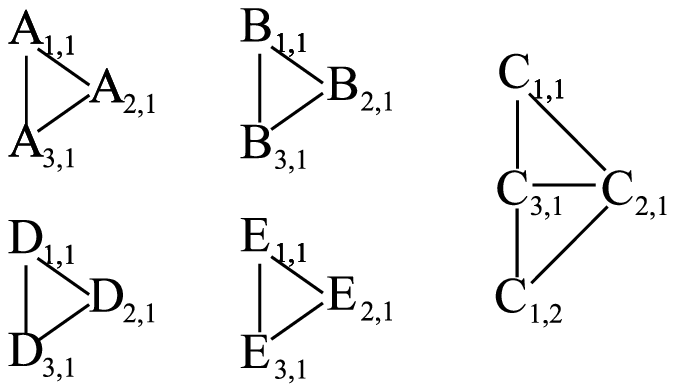}
  \end{minipage}
  \caption{The same alignment as in Fig. \ref{fig:sequences} and
    \ref{fig:sequences2} represented with the graph model.}
  \label{fig:graph1}
\end{figure}

The alignment is finally represented as a set of factor graphs. This
graph $G=(V, E)$ with $V$ the set of vertices where each vertex is on
factor of the alignment, and $E$ the set of edges between vertices
where an edge denotes the possibility for two common factors to be
aligned together in the same block.  The vertices of this graph are
numbered. That is to say a factor $X$, appearing for the $j^{th}$ time
in the $i^{th}$ sequence gives a vertex labeled $X_{i, j}$. Figure
\ref{fig:graph1} represent such a graph.

The constraints between edges that represent the conflicts between
factors have to be represented. In this way a second graph $G'$ is
constructed from the previous one. This new graph is isomorphic to the
previous one, such that the vertices of this new graph are created
from the edges of the previous one. From this new graph, a new set of
edges, that represent the conflicts between factors, is created.

\begin{figure}[htb]
\begin{minipage}[b]{1.0\linewidth}
\centering 
\includegraphics[width=0.7\textwidth]{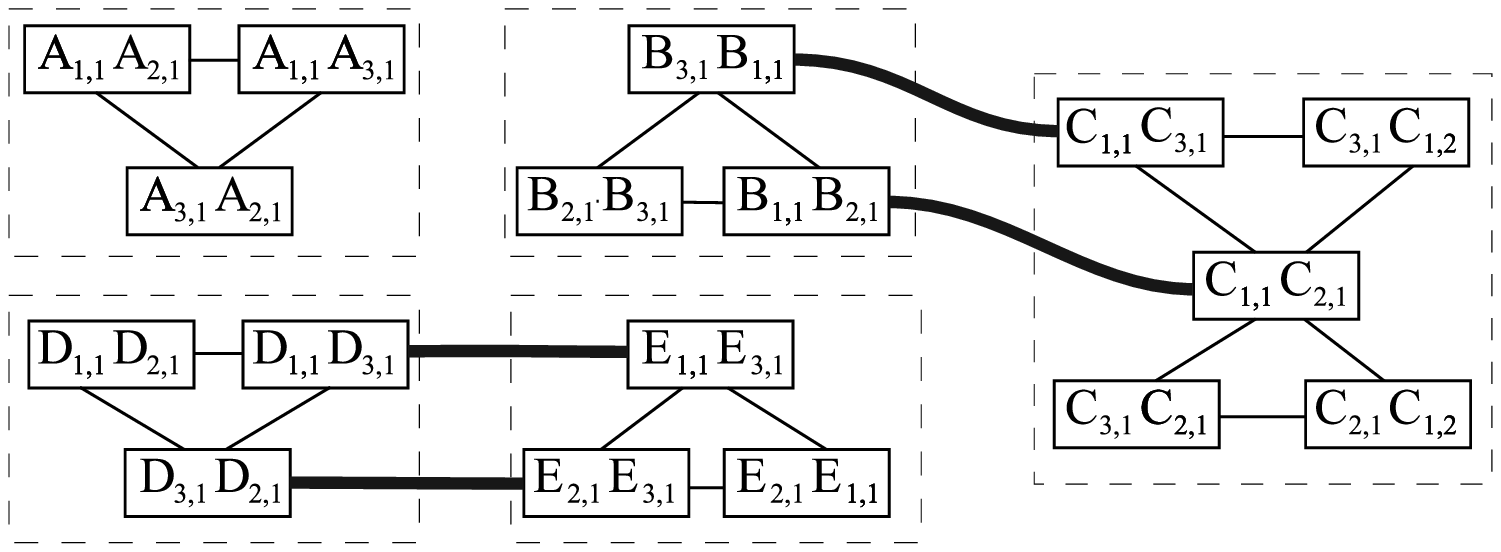}
\end{minipage}
\caption{This graph $G'$ is isomorphic to $G$. It is called
  \emph{conflicts graph}. It helps with the representation of
  conflicts between factors. The light edges represent the share of a
  common factor between two graphs. The Heavy edges represent the
  conflicts.}
  \label{fig:graph2} ()
\end{figure}

As it can be observed in Fig.  \ref{fig:graph2} these new vertices
have two kind of edges: one kind of edges when two vertices share a
common factor, and one kind of edge to represent the conflicts.

In the following algorithm, the graph $G$ is considered to be the
environment the ant will use. $G'$ is only used to determined
conflicts.

\subsection{The Algorithm}
\index{Multiple Sequence Alignment!Ant-based Systems|(}
\index{Ant-based Systems!Applications!Bioinformatics|see{Multiple Sequence Alignment}}

The general scheme of the behavior of the ant colony based system
follows these rules:
\begin{itemize}
\item Ants perform walks into the research graph.
\item During these walks they lay phe\-ro\-mones down the edges they
  cross.
\item They are attracted by the phe\-ro\-mone trails already laid down
  the environment.
\item Constrained areas of the graph are repulsive for ants.
\end{itemize}

Like has been said in the Sect. \ref{sec:approach}, the first step of
our approach is to characterize the solutions we are looking for.

\subsubsection{Formulation of a Solution}

This question is delicate because it returns to the nature of the
problem.  Indeed, multiple alignment is not only one fixed method.
Many biological considerations come into play. Nothing guarantees
besides that one is able to provide a coherent response without solids
knowledge in molecular biology.

Moreover, biologist themselves do not agree whether or not a given
alignment is a good one. Popular functions like the sum of pairs
\cite{Altschul89} are widely debated. In consequence, the formulation
of an evaluation function seems doubtful. It doesn't seem to be so
problematic since the method used doesn't evaluate solutions. Moreover
ants don't construct solutions individually.

Nevertheless, it is known that final alignments should be composed of
blocks without conflicts. The simplest and relatively natural idea is
to maximize the number of blocks formed without conflicts. But this
goal is not founded on any biological consideration. Such a goal can
probably draw aside from the solutions biologically better than those
required. An interesting property can be observed without scientific
considerations. Indeed starting from the factors one can evaluate the
relative distance between 2 blocks compared to the set the factors.
This information is simple and gives an advantage to edges between
close factors, ``geographically'' speaking.

Solutions in this model are carried out from the exploration of the
system composed of ant colonies. High valuable edges in the
environment end loaded with a large quantities of the phe\-ro\-mones.
The most heavily loaded edges belong to the solution. These edges link
vertices that represent parts of the blocks of the solution.

Since ants only perform walks in the environment, they do not consider
any block formation nor try to minimize any conflict. From a general
point of view, a non conflicted set of blocks rise from the graph.

\subsubsection{Phe\-ro\-mone Lay Down}

Let $\tau_{ij}$ be the quantity of phe\-ro\-mone present on the edge
$(i,j)$ of graph $G$ which connects the vertices $i$ and $j$. If
$\tau_{ij}(t)$ is the quantity of phe\-ro\-mone present on the edge
$(i,j)$ at moment $t$, then $\Delta\tau_{ij}$ is the quantity of
phe\-ro\-mone to be added to the total quantity on the edge at moment
$t+1$. So:
\begin{equation}
 \tau_{ij}(t+1)=(1-\rho).\tau_{ij}(t) + \Delta\tau_{ij}
\end{equation} 

$\rho$ represents the evaporation rate of the phe\-ro\-mones. Indeed,
the modeling of the evaporation (like natural phe\-ro\-mones) is
useful because it makes it possible to control the importance of the
produced effect. In practice, the control of this evaporation makes it
possible to limit the risks of premature convergence toward a local
minimum.

The quantity of phe\-ro\-mone $\delta\tau_{ij}$ added on the edge
$(i,j)$ is the sum of the phe\-ro\-mone deposited by all the ants
crossing the edge $(i,j)$ with the new step of time. If $m$ ants use
the edge $(i,j)$, then:

\begin{equation} 
\Delta\tau_{ij} = \sum_{k=1}^{m}{\Delta\tau_{ij}^k}
\end{equation}

The volume of phe\-ro\-mone deposited on each passage of an ant is a
constant value $Q$.

\subsubsection{Feedback Loops}

In a general scheme, negative feedback loops prevent the system from
continually increasing or decreasing to critical limits. In our
system, constraints between conflicted edges are repulsive for ants.
These negative loops aim at maintaining bounded quantities of
phe\-ro\-mones. Let consider an ant going over the edge $(i,j)$ of the
graph $G$. The information on the isomorphic graph $G'$ returns that
$(i,j)$ is in conflict with edges $(k,l)$ and $(m,n)$. Then the
quantities of phe\-ro\-mone will be modified consequently.
$\delta\tau_{ij}=\delta\tau_{ij}+q$ for the normal deposit of
phe\-ro\-mone and $\delta\tau_{kl}=\delta\tau_{kl}-q$ and
$\delta\tau_{mn}=\delta\tau_{mn}-q$ to represent negative feedback on
the edges in conflict with $(i,j)$.

\subsubsection{Transition Rule}

When an ant is on vertex $i$, the choice of the next vertex to be
visited must be carried out in a random way according to a definite
probability rule.

According to the method classically proposed in ant algorithms, the
choose of a next vertex to visit is influenced by 2 parameters. First,
a local heuristic is made with the local information that is the
relative distance between the factors. The other parameter is
representative of the stygmergic behavior of the system. It is the
quantity of phe\-ro\-mone deposited.

\par{Note:} Interaction between positive and negative phe\-ro\-mones
can lead on some edges to an overall negative value of phe\-ro\-mone.
Thus, phe\-ro\-mones quantities need normalization before the random
draw is made. Let $max$ be an upper bound value set to the largest
quantity of phe\-ro\-mones on the neighborhood of the current vertex.
The quantity of phe\-ro\-mone $\tau_{ij}$ between edge $i$ and $j$ is
normalized as $\tau_{ij} = max - \tau_{ij}$.

The function $neighbors(i)$ returns the list of vertices next to $i$.
The choice of the next vertex will be carried out in this list of
successors. The probability for an ant being on vertex $i$, to go on
$j$ ($j$ belonging to the list of successors) is:

\begin{equation} P(ij)=\frac{[
\frac{1}{max-\tau_{ij}}]^\alpha \times [ \frac{1}{d_{ij}
}]^\beta} {\sum_{s \in neighbors(i))} [
\frac{1}{max-\tau_{is}}]^\alpha \times [ \frac{1}{d_{is}}]^\beta}
\end{equation}

In this equation, the parameters $\alpha$ and $\beta$ make it possible
more or less to attach importance to the quantities of phe\-ro\-mone
compared to the relative distances between the factors.

\index{Multiple Sequence Alignment!Ant-based Systems|)}

\subsection{Results}
\label{sec:msa_res}

In the examples below, for a better readability, the factors are
reduced to letters. A sequence is only represented with its factors.
For example, {\tt ABBC} is a sequence containing 4 factors, one factor
labeled 'A', two labeled 'B' and 1 named 'C'.

Let's see some little alignments that require a particular attention.
In the tables, the first column represents the original alignment. The
other columns represent possible solutions. The last line of the table
show the average number of time solutions are chosen over 10 trials.

\begin{table}
\centering
\caption{}\label{t1}

{\tt
\begin{tabular}{l  c   c   c}
\hline\noalign{\smallskip}
AB~~ & ~~AB-~~ & ~~-AB~~ & ~~AB~~ \\ 
AB~~ & ~~AB-~~ & ~~-AB~~ & ~~AB~~\\ 
BA~~ & ~~-BA~~ & ~~BA-~~ & ~~BA~~ \\ 
\hline
& $5$ & $5$ & $0$ \\
\hline\noalign{\smallskip}
\end{tabular}
}
\end{table}

The choice, in the table \ref{t1} must be made between the 'A' factors
and the 'B' factors to determine which block will be complete. The
algorithm cannot make a clear choice between the 2 first solutions.

\begin{table}
\centering
\caption{}\label{t2}
{\tt
\begin{tabular}{l  c  c  c }
\hline\noalign{\smallskip}
AB~~  & ~~AB-~~ & ~~--AB~~ & ~~--AB-~~\\ 
AA~~  & ~~A-A~~ & ~~A-A-~~ & ~~--A-A~~\\ 
ABA~~ & ~~ABA~~ & ~~ABA-~~ & ~~ABA--~~\\ 
\hline & $10$ & $0$ & $0$\\
\hline\noalign{\smallskip}
\end{tabular}
}
\end{table}

In the table \ref{t2}, the choice of 3 blocks is wanted. The
difficulty for the method is to discover the 'B' block inside all the
'A' factors.

\begin{table}
\centering
\caption{}\label{t3}

{\tt
\begin{tabular}{l c  c  c}
\hline\noalign{\smallskip}
DAZZZZ~~ & ~~DA-----ZZZZ~~ & ~~DAZZZZ-----~~ & ~~-----DAZZZZ~~\\ 
ACGSTD~~ & ~~-ACGSTD----~~ & ~~-A----CGSTD~~ & ~~ACGSTD-----~~\\ 
ACGSTD~~ & ~~-ACGSTD----~~ & ~~-A----CGSTD~~ & ~~ACGSTD-----~~\\ 
ACGSTD~~ & ~~-ACGSTD----~~ & ~~-A----CGSTD~~ & ~~ACGSTD-----~~\\ 
\hline & $0$ & $10$ & $0$\\
\hline\noalign{\smallskip}
\end{tabular}
}

\end{table}

In the table \ref{t3}, the 'A' factor and the 'D' factor on the first
sequence have the same conflicts. The choice will be made thanks to
the relative distance between the factors. Here, the rest of he 'D'
factor are far from the 'D' factor of the sequence 1, so it will not
be chosen.

\begin{table}
\centering
\caption{}\label{t4}
{\tt
 \begin{tabular}{l c  c  c}
\hline\noalign{\smallskip}
BABAB~~   & ~~B-ABA-B~~  & ~~-BABA-B~~ & ~~-BAB-AB~~\\ 
BBABAAB~~ & ~~BBABAAB~~  & ~~BBABAAB~~ & ~~BBABAAB~~\\ 
BABAB~~   & ~~B-ABA-B~~  & ~~-BABA-B~~ & ~~-BAB-AB~~\\
\hline & $0$ & $5$ & $5$ \\
\hline\noalign{\smallskip}
\end{tabular}
}

\end{table}

The goal, in table \ref{t4} tests the aggregation capacity of the
method. It is the concept of "meta-block" that we want to highlight.
Here, the first block of 'B' factors can be align with the first 'B'
factor of the second sequence or with the second factor 'B' on the
same sequence. The algorithm will chose the second option because
there is a meta-block at this position, i.e. two or more blocks where
each factor on each sequence follow in the same order.

\index{Multiple Sequence Alignment|)}

\vfill
\pagebreak

\section{Natural Language Processing}
\label{sec:nlp}
\index{Natural Language Processing|(}

\subsection{Description of the Problem}
\label{sec:nlp_pb}

\begin{figure}[htb]
\begin{minipage}[b]{1.0\linewidth}
  \centering 
  \includegraphics[width=0.8\textwidth]{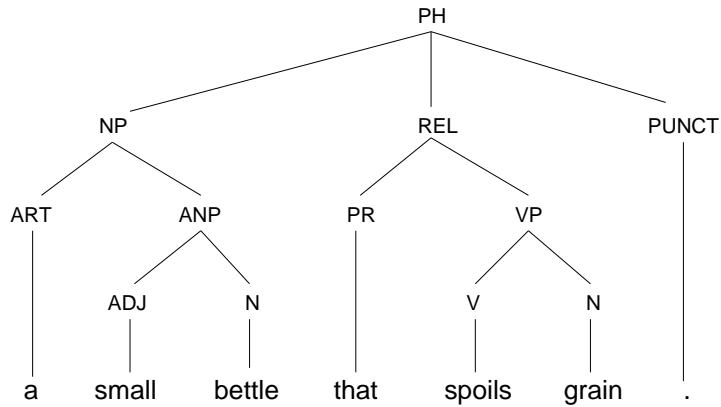}
\end{minipage}
\caption{Morphosyntactic tree obtained from a lexical and grammatical
  analysis of the text. ART denotes article, PUNCT the punctuation, V
  stands for verb, N for noun... \label{fig:morphosyntactic-tree}}
\end{figure}

Understanding a text requires the comparison of the various meanings
of any polysemous word with the context of its use.  But, the context
is precisely defined by the words themselves with all their meanings
and their status in the text (verb, noun, adjective...).  Given their
status, there exist some grammatical relations between words.  These
relations can be represented by a tree-like structure called a
morpho-syntactic analysis tree.  Such a tree is represented on Fig.
\ref{fig:morphosyntactic-tree}.

This structure represents the syntactical relations between words, but
in no way their semantics relations.  In order to represent this new
relational structure, we consider words as the basic compounds of an
interaction network which implicit dynamics reveals the pregnancy of
each meaning associated to any polysemous word.  If we refer to the
most commonly shared definition of a {\em complex system}, it states
that it is a {\em network of entities which interactions lead to the
  emergence of structures that can be identified as high-level
  organizations}.  The action of one entity may affect subsequent
actions of other entities in the network, so that {\em the action of
  the whole is more than the simple sum of the actions of its
  parts.}\footnote{\cite{AL96} {\em Why do we need artificial life ?}
  page 305.}.  The {\em actions} in our context correspond to the {\em
  meanings} of the words constituting the text, and the sum of the
actions results in the global meaning of the text, which is, for sure,
much more than the simple
sum of the meanings of the words.\\

The addressed problem is twofold.  On the one hand, we have to find a
way of expressing words meanings interactions, and on the other hand,
we have to conceive a system able to bring to the fore potential
meanings for the whole text in order to help an expert for raising
ambiguities due to polysemy.  This problem is known as Word Sense
Disambiguation (WSD).  \index{Word Sense Disambiguation (WSD)} 
A more detailed description of the proposed solution can be found in 
\cite{Lafourcade_Guinand_2006}.

\subsection{Models and Tools}
\label{sec:nlp_model_tools}

\begin{figure}[htb]
 \begin{minipage}[b]{.48\linewidth}
   \centering
   \includegraphics[width=4cm]{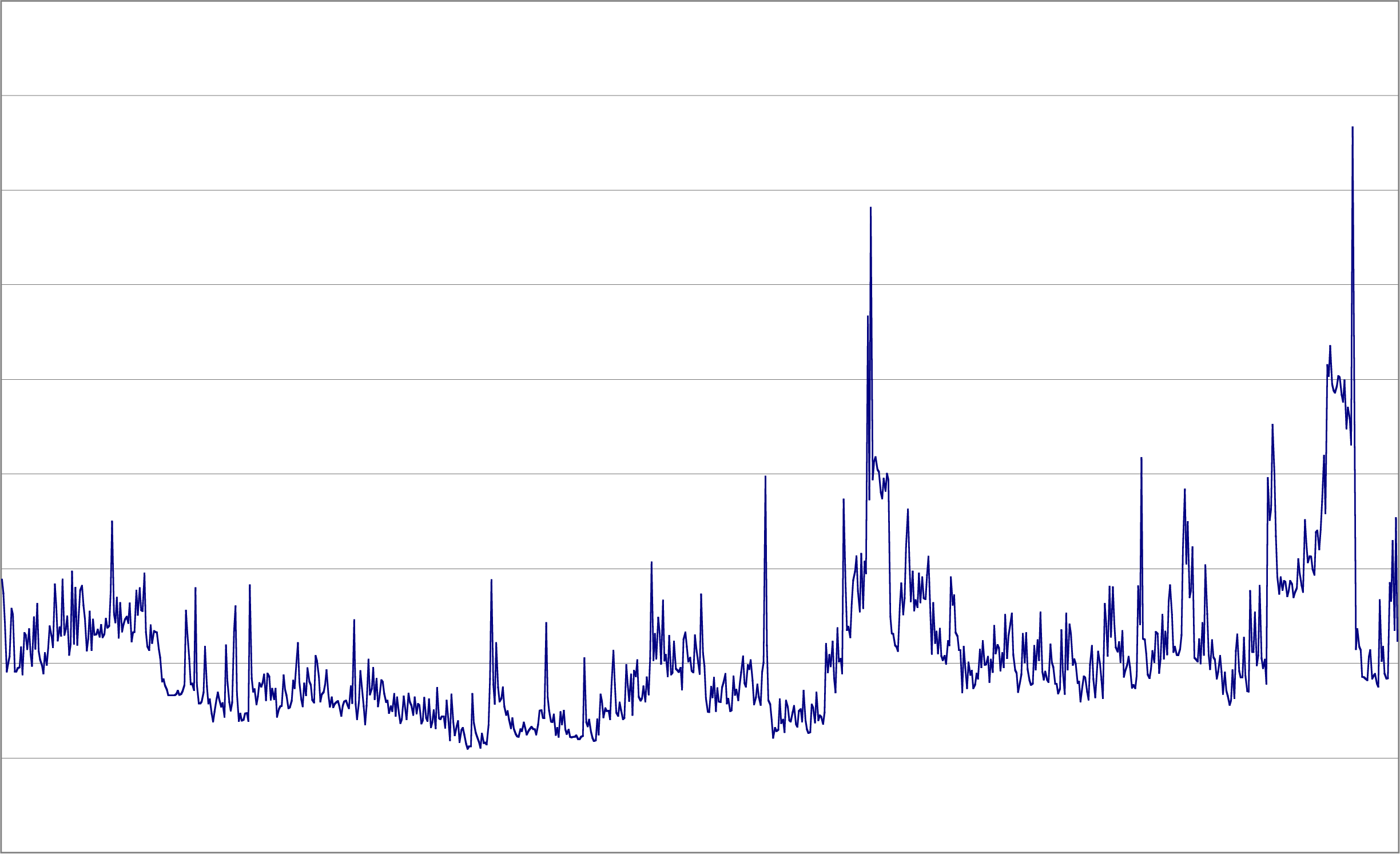}
   \centerline{{\em profit} (profit).\label{fig:profit}}\medskip
 \end{minipage}
 \hfill
 \begin{minipage}[b]{0.48\linewidth}
   \centering
   \includegraphics[width=4cm]{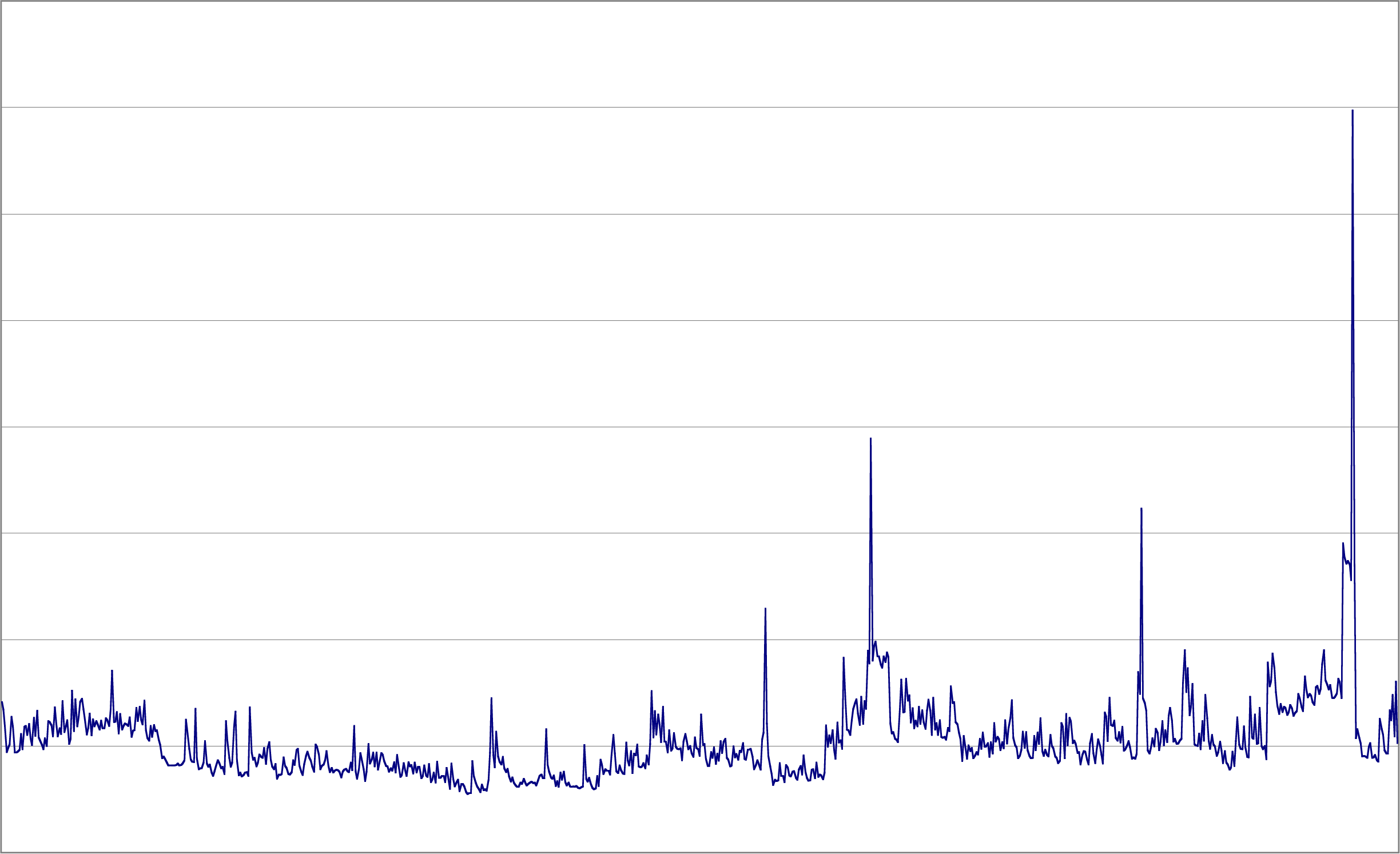}
   \centerline{{\em benefice} (benefit).\label{fig:benefice}}\medskip
 \end{minipage}
 \caption{Conceptual vectors of two terms (profit and benefit). 
          In French, we can observe that {\em profit} is more polysemous than
	  {\em benefice}. Moreover, from the graphical representation,
	  it appears that both terms are closely related.}
 \label{fig:vectors}
\end{figure}

Thematic aspects of textual segments (documents, paragraphs, syntagms,
etc.) can be represented by conceptual vectors.  From a set of
elementary notions, concepts, it is possible to build vectors
(conceptual vectors) and to associate them to lexical
items\footnote{Lexical items are words or expressions that constitute
  lexical entries. For instance, \em{car} or \em{white ant} are
  lexical items}.  Polysemous words combine the different vectors
corresponding to the different meanings (Fig. \ref{fig:vectors} gives
an illustration of conceptual vectors).

This vector approach is based on well known mathematical properties,
it is thus possible to undertake well founded formal manipulations
attached to reasonable linguistic interpretations.  Concepts are
defined from a thesaurus (in the prototype applied to French,
\cite{LAR92} has been chosen where 873 concepts are identified).  The
building of conceptual vectors is a research domain in itself and is
out of the scope of this brief example, so in the following, we
consider that each meaning of a word is attached to a conceptual
vector.

In order to build the interaction network we need a way of deciding
whether two terms are thematically related or not.  For that purpose,
we define two measures: $Sim(A,B)$ a {\em similarity measure} between
two vectors $A$ and $B$ that can be expressed as the scalar product of
both vectors divided by the product of their norm, and the
\emph{angular distance} $D_{A}$ between two vectors $A$ and $B$
($D_{A}(A,B) = \arccos(Sim(A,B))$) Intuitively, this latter function
constitutes an evaluation of the \emph{thematic proximity} and is the
measure of the angle between the two vectors.  We would generally
consider that, for a distance $D_{A}(A,B) \leq \frac{\pi}{4}$, (i.e.
less than 45 degrees) A and B are thematically close and share many
concepts.  For $D_{A}(A,B) \geq \frac{\pi}{4}$, the thematic proximity
between A and B would be considered as loose.
Around $\frac{\pi}{2}$, they have no relation.\\

The structures associated to solutions for our problem can be defined
as a path or a set of paths.  Each path correspond to a so-called {\em
  interpretation trail}.  The fact that within a sentence several
meanings may co-exist entails the possibility for several
interpretation trails and thus of several distinct paths.

\subsection{Ant-based Method}
\label{sec:method}
\index{Natural Language Processing!Ant-based Systems|(}
\index{Ant-based Systems!Applications!Natural Language Processing|(}

Our method for WSD relies on both kind of stigmergy.  Sign-based
stigmergy plays a role in ant behaviors.  Sematectonic stigmergy is
used for modifying nodes characteristics and for creating new paths
between vertices.  In the sequel, these new paths will be called
bridges.

The "binary bridge" is an experiment developed by
\cite{Pasteels_et_al_1987}.  As reported in
\cite{Bonabeau_Dorigo_Theraulaz_1999} {\em in this experiment, a food
  source is separated from the nest by a bridge with two equally long
  branches A and B.}  Initially, both paths are visited and after some
iterations, one path is selected by the ants, whereas the second,
although as good as the first one, is deserted.  This experiment
interests us for two reasons.  It first shows that ants have the
ability of organizing themselves in order to determine a global
solution from local interactions, thus, it is likely to obtain an
emergent solution for a problem submitted to an ant-based method.
This point is crucial for our approach, since we expect the emergence
of a meaning for the analyzed text.  But, the experiment also shows
the inability of such method, in its classical formulation, to provide
a set of simultaneous and distinct solutions instead of only one at a
time.  As these methods are based on the reinforcement of the current
best solution, they are not directly suitable for our situation.
Indeed, if several meanings are possible for a text, all these
meanings should emerge.  That was the main motivation for introducing
color as a colony characteristic.

The computational problem is twofold.  Indeed the meanings are not
strictly speaking active entities. In order to ensure the interactions
of the meanings of the whole text, an active framework made of
"meaning transporters" must be supplied to the text.  These
"transporters" are intended to allow the interactions between meanings
of text elements.  They have to be both light (because of their
possible large number) and independent (word meanings are intrinsic
values).  Moreover, when some meanings stemmed from different words
are compatible ({\em engaged} with {\em job} for instance), the system
has to keep a trace of this fact.\\
\index{Dynamic Graphs}
This was the original motivation for us to consider ant-based complex
systems as described in Sect. \ref{sec:approach}.  A similar idea
already existed in \cite{HOF95} with the COPYCAT project, in which the
environment by itself contributes to solution computation and is
modified by an agent population where roles and motivations varies.
We retain here some aspects mentioned in \cite{Gales92}, that we
consider as being crucial: (1) mutual information or semantic
proximity is one key factor for lexical activation, (2) the syntactic
structure of the text can be used to guide information propagation,
(3) conceptual bridges can be dynamically constructed (or deleted) and
could lead to \emph{catastrophic events} (in the spirit of
\cite{THOM}).  These bridges are the instrumental part allowing
mutual-information exchange beyond locality horizons.  Finally, as
pointed by \cite{HOF95}, biased randomization (which doesn't mean
chaos) plays a major role in the model.

In order to build several structures corresponding to competing
meanings, we consider several colonies with the color characteristic.

\subsection{Environment}
\label{sec:nlp_env}

The underlying structure of the environment is the morphosyntactic
analysis tree of the text to be analyzed.  Each content word is a
node. This node has as many children in the tree as senses.  To each
child associated to a sense corresponds a unique color: the conceptual
vector of the sense.  A child is also a {\em nest} and all children of
a node associated to a content word are {\em competing nests}.  An ant
can walk through graph edges and, under some circumstances, can build
new ones (called bridges).  Each node contains the following
attributes beside the morphosyntactic information computed by the
analyzer: (1) a resource level $R$, and (2) a conceptual vector $V$.
Each edge contains (1) a phe\-ro\-mone level.  The main purpose of
phe\-ro\-mone is to evaluate how popular a given edge is.  The
environment by itself is evolving in various aspects:

\begin{enumerate}
\item the conceptual vector of a node is slightly modified each time a
  new ant arrives.  Only vectors of nests are invariant (they cannot
  be modified).  A nest node is initialized with the conceptual vector
  of its word sense, other nodes with the null vector.
\item resources tend to be redistributed toward and between nests
  which \emph{reinvest} them in ants production.  Nodes have an
  initial amount of resources of 1.
\item the phe\-ro\-mone level of edges are modified by ant moves.  The
  evaporation rate $\delta$ ensures that with time phe\-ro\-mone level
  tends to decrease toward zero if no ant are passing through.  Only
  bridges (edges created by ants) would disappear if their
  phe\-ro\-mone level reaches zero.
\end{enumerate}

The environment has an impact on an ant and in return ants
continuously modify the environment.  The results of a simulation run
are decoded thanks to the phe\-ro\-mone level of bridges and the
resource level of nests.

\subsection{Nests, Ant Life and Death}
\label{sec:colored_ants}

A nest (word sense) has some resources which are used for producing
new ants.  At each cycle, among the set of nests having the same
parent node (content word), only one is allowed to produced a new ant.
The color of this ant is the one of the selected nest.  In all
generality, a content word has $n$ children (nests), and the nest
chosen for producing the next ant is probabilistically selected
according to the level of resources.  There is a cost $\epsilon$ for
producing an ant, which is deducted from the nest resources. Resource
levels of nests are modified by ants.  The probability of producing an
ant, is related to a sigmoid function applied to the resource level of
the nest.  The definition of this function ensures that a nest has
always the possibility to produce a new ant although the chances are
low when the node is inhibited (resources below zero).  A nest can
still borrow resources and thus a word meaning has still a chance to
express itself even if the environment is very unfriendly.

The ant cost can be related to the ant life span $\lambda$ which is
the number of cycles the ant can forage before dying.  When an ant
dies, it gives back all the resources it carries plus its cost, to the
currently visited node.  This approach leads to a very important
property of the system, that the total level of resources is constant.
The resources can be unevenly distributed among nodes and ants and
this distribution changes over time, sometimes leading to some
stabilization and sometimes leading to periodic configurations.  This
is this \emph{transfer of resources} that reflects the lexical
selection, through word senses activation and inhibition.

The ant population (precisely the color distribution) is then evolving
in a different way of classical approaches where ants are all similar
and their number fixed in advance.  However, at any time (greater than
$\lambda$), the environment contains at most $\lambda$ ants that have
been produced by the nests of a given content word.  It means that the
global ant population size depends on the number of content words of
the text to be analyzed, but not on the number of word meanings.  To
our views, this is a very strong point that reflects the fact some
meanings will express more than others, and that, for a very polysemic
word, the ant struggle will be intense.  A monosemic word will often
serve as a pivot to other meanings.  Moreover, this characteristic
allows us to evaluate the computing requirements needed for computing
the analysis of a given text since the number of ants depends only on
the number of words.

\subsection{Ant Population}

An ant has only one motivation: foraging and bringing back resources
to its nest.  To this purpose, an ant has two kinds of behavior
(called modes), (1) searching and foraging and (2) returning resources
back to the nest.  An ant $a$ has a resource storage capacity $R(a)
\in [0, 1]$.  At each cycle, the ant will decide between both modes as
a linear function of its storage.  For example, if the $R(a) = 0.75$,
there is a $75\%$ chance that this ant $a$ is in \emph{bringing back}
mode.

Each time an ant visits a (non-nest) node, it modifies the node color
by adding a small amount of its own color.  This modification of the
environment is one factor of the sematectonic stigmergy previously
mentioned and is the means for an ant to find its way back home.  The
new value of the color is computed as follows: $C(N) = C(N) + \alpha
C(a)$ with $0 < \alpha < 1$.  In our application, colors are
conceptual vectors and the ``+'' operation is a normalized vector
addition ($V(N) = V(N) + \alpha V(a)$).  We found heuristically, that
$\alpha=1/\lambda$ constitutes a good trade-off between a static and a
versatile environment.

\subsection{Searching Behavior}

Given a node $N_i$. $N_j$ is a neighbor of $N_i$ if and only if there
exists an edge $E_{ij}$ linking both nodes.  A node $N_i$ is
characterized by a resource level noted as $R(N_i)$.  An edge $E_{ij}$
is characterized by a phe\-ro\-mone level noted as $Ph(E_{ij})$.  A
searching ant will move according to the resource level of each
adjacent node (its own nest excepted) and to the level of
phe\-ro\-mones of the outgoing edges.  More precisely an attraction
value is computed for each neighbor.  This value is proportional to
the resource level and inversely proportional to the phe\-ro\-mone
level.

\begin{equation}
\displaystyle
\mbox{attract}_S(N_x) = \frac{\max \left(\mbox{R}(N_x),\eta \right)}
	{\mbox{Ph}(E_{ix}) + 1}
\end{equation}

Where $\eta$ is a tiny constant avoiding null values for attraction.
The motivation for considering an attraction value proportional to the
inverse of the phe\-ro\-mone level is to encourage ants to move to non
visited parts of the graph.  If an ant is at node $N_i$ with $p$
neighbors $N_k (k = 1 \cdots p)$, the probability $P_S(N_x)$ for this
ant to choose node $N_x$ in \emph{searching} mode is:

\begin{equation}
\displaystyle
P_S(N_{x}) = \frac{\mbox{attract}_S(N_x)}
	{\sum_{1\leq j \leq p} \mbox{attract}_S(N_j)}
\end{equation}

Then, if all neighbors of a node $N_i$ have the same level of
resources (including zero), then the probability for an ant visiting
$N_i$ to move to a neighbor $N_x$ depends only on the phe\-ro\-mone
level of the edge $E_{ix}$.

%
An ant is attracted by node with a large supply of resources, and will
take as much as it can hold (possibly all node resources).  A depleted
node does not attract searching ants.  The principle here, is a simple
greedy algorithm.

\subsection{Bringing Back Behavior}

When an ant has found enough resources, it tends to bring them back to
its nest.  The ant will try to find its way back thanks to the color
trail left back during previous moves.  This trail could have been
reinforced by ants of the same color, or inversely blurred by ants of
other colors.

An ant $a$ returning back and visiting $N_i$ will move according to
the color similarity of each neighboring node $N_x$ with its own color
and according to the level of phe\-ro\-mones of the outgoing edges.
More precisely an attraction value is computed for each neighbor.
This value is proportional to the similarity of colors and to the
phe\-ro\-mone
level:\\

$\mbox{attract}_R(N_x) = \max
(\mbox{sim}(\mbox{colorOf}(N_x),\mbox{colorOf}(a)),\eta) \times
(\mbox{Ph}(E_{ix}) + 1)$
~\\

Where $\eta$ is a tiny constant avoiding null values for attraction.

If an ant is at node $N_i$ with $p$ neighbors $N_k (k = 1 \cdots p)$,
the probability $P_B(N_x)$ for this ant to choose node $N_x$ in
\emph{returning} mode is:

\begin{equation}
\displaystyle
P_R(N_{x}) = \frac{\mbox{attract}_R(N_x)}
	{\sum_{1\leq j \leq p} \mbox{attract}_R(N_j)}
\end{equation}

%

All considered nodes are those connected by edges in the graph.  Thus,
the syntactic relations, projected into geometric neighborhood on the
tree, dictate constraints on the ant possible moves.  However, when an
ant is at a friendly nest, it can create a shortcut (called a
\emph{bridge}) directly to its home nest.  That way, the graph is
modified and this new arc can be used by other ants.  These arcs are
evanescent and might disappear when the phe\-ro\-mone level becomes
null.

From an ant point of view, there are two kinds of nests: friend and
foe.  Foe nests correspond to alternative word senses and ants stemmed
from these nests are competing for resources.  Friendly nests are all
nests of other words.  Friends can fool ants by inciting them to give
resource.  Foe nests instead are eligible as resource sources, that is
to say an ant can steal resources from an enemy nest as soon as the
resource level of the latter is positive.

\index{Natural Language Processing!Ant-based Systems|)}
\index{Ant-based Systems!Applications!Natural Language Processing|)}

\subsection{Results}
\label{sec:nlp_res}

The evaluation of our model in terms of linguistic analysis is by
itself challenging.  To have a larger scale assessment of our system,
we prefer to evaluate it through a Word Sense Disambiguation task
(WSD).

A set of 100 small texts have been constituted and each term (noun,
adjective, adverb and verb) has been manually tagged.  A tag is a term
that names one particular meaning.  For example, the term \emph{bank}
could be annotated as \emph{bank/ri\-ver}, \emph{bank/\-mo\-ney
  ins\-ti\-tu\-tion} or \emph{bank/building} assuming we restrict
ourselves to three meanings.  In the conceptual vector database, each
word meaning is associated to at least one tag (in the spirit of
\cite{Jalabert_Lafourcade_2002}).  Using tag is generally much easier
than sense number especially for human annotators.

The basic procedure is quite straightforward.  The unannotated text is
submitted to the system which annotates each term with the guessed
meaning.  This output is compared to the human annotated text.  For a
given term, the annotation available to the human annotator are those
provided by the conceptual vector lexicon (i.e. for bank the human
annotator should choose between \emph{bank/ri\-ver},
\emph{bank/\-mo\-ney ins\-ti\-tu\-tion} or \emph{bank/building}).  It
is allowed for the human annotator to add several tags, in case
several meanings are equally acceptable.  For instance, we can have
\emph{The frigate/\{modern ship/ancient ship\} sunk in the harbor.},
indicating that both meanings are acceptable, but excluding
\emph{frigate/bird}.  Thereafter, we call \emph{gold standard} the
annotated text.  We should note that only annotated words of the gold
standard are target words and used for the scoring.

When the system annotates a text, it tags the term with all meanings
which activation level is above 0.  That is to say that inhibited
meanings are ignored.  The system associates to each tag the
activation level in percent.  Suppose, we have in the sentence
\emph{The frigate sunk in the harbor.} an activation level of
respectively $1.5$ , $1$ and $-0.2$ for respectively
\emph{frigate/modern ship}, \emph{frigate/ancient ship} and
\emph{frigate/bird}.  Then, the output produced by the system is:

\begin{quote}
\emph{The frigate/\{modern ship:0.6/ancient ship:0.4\}}.
\end{quote}

Precisely, we have conducted two experiments with two different
ranking methods.

\begin{quote}
  A \emph{Fine Grained} approach, for which only the first best
  meaning proposed by the system is chosen.  If the meaning is one of
  the gold standard tag, the answer is considered as valid and the
  system scores $1$.  Otherwise, it is considered as erroneous and the
  system scores $0$.

  A \emph{Coarse Grained} approach, more lenient, gives room to
  closely related meanings.  If the first meaning is the good one,
  then the system scores $1$.  Otherwise, the system scores the
  percent value of a good answer if present.  For example, say the
  system mixed up completely and produced:
\begin{quote}
\emph{The frigate/\{bird:0.8/ancient ship:0.2\}}.
\end{quote}

the system still gets a $0.2$ score.
\end{quote}

\begin{table}
\centering
\caption{}
\begin{tabular}{l l l l l l} 
\hline\noalign{\smallskip}
{\bf Scoring scheme} &{\bf All terms} &{\bf Nouns} 	& {\bf Adjectives} 	& {\bf Verbs}  	& {\bf Adverbs} \\ \hline
Fine Grain Scoring   &	   0.68	    &  	0.76   	& 	0.78      	&	0.61	&	0.85 \\
Coarse Grain Scoring &	   0.85	    &	0.88   	& 	0.9      	&	0.72	&	0.94 \\
\hline\noalign{\smallskip}
 \end{tabular}
\end{table}

These results compare quite favorably to other WSD systems as
evaluated in SENSEVAL campaign \cite{Senseval_2000}.  However, our
experiment is applied to French which figures are not available in
Senseval-2 \cite{Senseval_2001}.

As expected, verbs are the most challenging as they are highly
polysemous with quite often subtle meanings.  Adverbs are on the
contrary quite easy to figure when polysemous.

We have manually analyzed failure cases.  Typically, in most cases the
information that allows a proper meaning selection are not of thematic
value.  Other criteria are more prevalent.  Lexical functions, like
hyperonymy (is-a) or meronymy (part-of) quite often play a major role.
Also, meaning frequency distribution can be relevant.  Very frequent
meanings can have a boost compared to rare ones (for example with a
proportional distribution of initial resources).  Only if the context
is strong, then could rare meanings emerge.

All those criteria were not modeled and included in our experiments.
However, the global architecture we propose is suitable to be extended
to ants of other \emph{caste} (following an idea developed by Bertelle
et al. in \cite{AAMAS_2002}).  In the prototype we have developed so
far, only one caste of ants exists, dealing with thematic information
under the form of conceptual vectors.  Some early assessments seem to
show that only with a semantic network restricted \emph{part-of} and
\emph{is-a} relations, a 10\% gain could be expected (roughly a of
gain 12\% and a lost of 2\%).

\index{Natural Language Processing|)}

\vfill
\pagebreak

\section{Analysis} 
\label{sec:analysis}
From the problems presented Sects. \ref{sec:msa} and \ref{sec:nlp} the
three main steps identified in Sect. \ref{sec:approach} can be argued.
 
 \subsection{Identification of the Structures}
\index{Graphs!Structures}

The first necessity  when using our approach is to clearly 
 identify  and define the shape of the desired structures in the environment 
 graph.  This definition must be clearly established when  modeling the 
 problem. In other words, the conception of a graph model for one particular 
 problem must be done in consideration of the necessary structures.  Finally, 
 the generated model must be practicable by the system and  must  permit the 
 production of such structure.
 
 \subsection{Choice of General Features to Achieve the Goal}

 Secondly, from the description of the wanted structures, a choice of
 general characteristics for nests, colonies and ants has to be made.
 These characteristics define general behaviors that lead the system
 to produce wanted structures. Let consider as an example two general
 groups of characteristics that lead to two different kind of
 structures:
 \begin{itemize}
 \item If a path or a set of paths is desired as solution to one
   problem, then, some nests must be defined in the graph. They are
   assigned a vertex of the search graph.  Ants are given one or more
   destinations vertices so as to perform walks between a nest and a
   destination. Thus, the system will produce paths between two
   vertices. The phe\-ro\-mone deposit can be performed when going
   back to the nest. A tabu list is used, its size depends heavily on
   the shape of the graph but it is generally equal to the size of the
   whole tour of the ant. The tabu list prevents ants form making
   loops in their paths.
 \item If sub-graphs or sets of vertices/edges are wanted, there is
   not one special vertex that represents the nest, indeed, ants are
   displayed on every vertices of the graph. So it is considered that
   each vertex of the graph is a nest.  More than one colony is used
   so as to define competitive behaviors between ants.
   \index{Colored Ants}
Ants are assigned a colony specific color and deposit colored
   phe\-ro\-mones at each edge cross. A tabu list is used but its
   purpose is to prevent ants from stagnating on a same couple of
   vertices, so it has a relatively small size.
 \end{itemize}
 This general settings define a global behavior of the system for
 common kinds of solution structures but are not enough to lead the
 system in a precise problem.

 \subsection{Relevant Parameters Tuning}
 
 Thirdly, after the definition of a structure and a general set of
 characteristics to solve one problem, a set of relevant parameters
 need to be identified. Modifications of these parameters bring
 behavioral changes to the system.  In effect, once the general
 process defined, these relevant parameters permit the adaptation to
 the precise problem and to the precise environment.  Finally the
 tuning of those parameters sounds quiet difficult and require
 attention.  As said by Theraulaz in \cite{Theraulaz_1997}, one of the
 {\em signatures} of self-organizing processes is their sensitivity to
 the values of some critical parameters. It has been observed from
 experiments that some parameters are probably linked there is
 probably a sort of ratio that is to be found between them. Another
 idea that these parameters are strongly related to the nature of the
 environment that is considered. One simple example of this
 correlation is the relative dependency between the number of ants and
 the number of elements (vertices and edges) of the graph. It is true
 that a bigger graph will require a bigger colony to explore it, but
 how much ant for how much element is an hard question and probably
 depends to other values like the average degree of each vertex (its
 connectivity).

 \subsection{An Example of this Analysis}

 The quite simple shortest path problem has been modeled. In this
 problem, the shortest path between two points of an environment is to
 be found. The environment is modeled with a graph where the start
 point and the end point are two particular vertices of this graph.

 The first step that is the definition of the desired structures is
 straightforward. One solution is one path in the graph starting from
 one of the two previously defined points and stopping at the other
 one.

 In the second step, the general characteristics are defined so as to
 create this path.  One ant colony is used. One nest is located on the
 start vertex.  Ants start their exploration from the nest and are
 looking for the end vertex. When the destination is found, they go
 back to the nest by laying phe\-ro\-mone down the edges of the paths
 they found.  They use a tabu list who's size is of the order of the
 constructed path's length.

 In the third step, experiments allowed the identification of four
 relevant parameters:

\begin{itemize}
\item The number of ants, that define the total number of ant agent in
  the colony. It seams to be highly related to the distribution of the
  edges in the graph.
\item The evaporation rate for the phe\-ro\-mone quantities. If a high
  evaporation value is set, it is difficult for ants influence each
  other with phe\-ro\-mones.  They will explore new parts of the graph
  better than exploiting already visited areas. If the evaporation is
  set to a lower value, phe\-ro\-mone trails evapo\-rate slow\-ly,
  they are stronger and bring ants to already visited paths.
\item The tabu list size defines the number of already visited edges
  that cannot be crossed again. This parameter prevents ants from
  making loops.  It would normally be set to the length of the
  constructed path, in other words, ants are not allowed to visit
  twice the same vertex.  Actually, this parameter needs sometimes to
  be set to a lower value depending on the shape of the graph.
\item The exploration threshold defines the sensitivity of the ants to
  the phe\-ro\-mone trails. It is usually used to find a balance
  between the intensification of phe\-ro\-mo\-ne trails and the
  diversification of the search. For one ant situated on vertex $i$,
  this parameter ($q_0$) influences the calculation of the transition
  rule. $q_0$ is the probability ($0 \leq q_0 \leq 1$) of choosing the
  most heavily loaded edge in phe\-ro\-mones among outgoing edges of
  the local vertex $i$. The normal transition rule
  (\ref{eq:transRule}) is chosen with probability $1-q_0$.

  When ants are too much attracted by phe\-ro\-mones they follow
  already defined trails and don't try to visit unexplored parts of
  the graph. On the contrary, if they don't pay enough attention to
  the phe\-ro\-mones, no stigmergy is available.
\end{itemize}

As an example, let's consider a torus graph composed of 375 nodes and
750 edges.  The shortest path in this graph has to be determined
between two defined vertices. The shortest path between those vertices
is 20 edges long.  A torus has no border.  Each vertex has exactly 4
outgoing edges.  Among experiments carried out the following parameter
set gives good results for his graph:
\begin{itemize}
\item Number of ants = 100.
\item Evaporation rate = 0.03. Note: the evaporation process is
  performed at each step. One step represents one vertex move for each
  ant of the colony.
\item Tabu list size = size of the length of the path.
\item Exploration threshold = 0.6.
\end{itemize}

\begin{center}
\begin{figure}[htb]
\begin{minipage}[b]{0.49\linewidth}
  \centering
  \includegraphics[width=\textwidth]{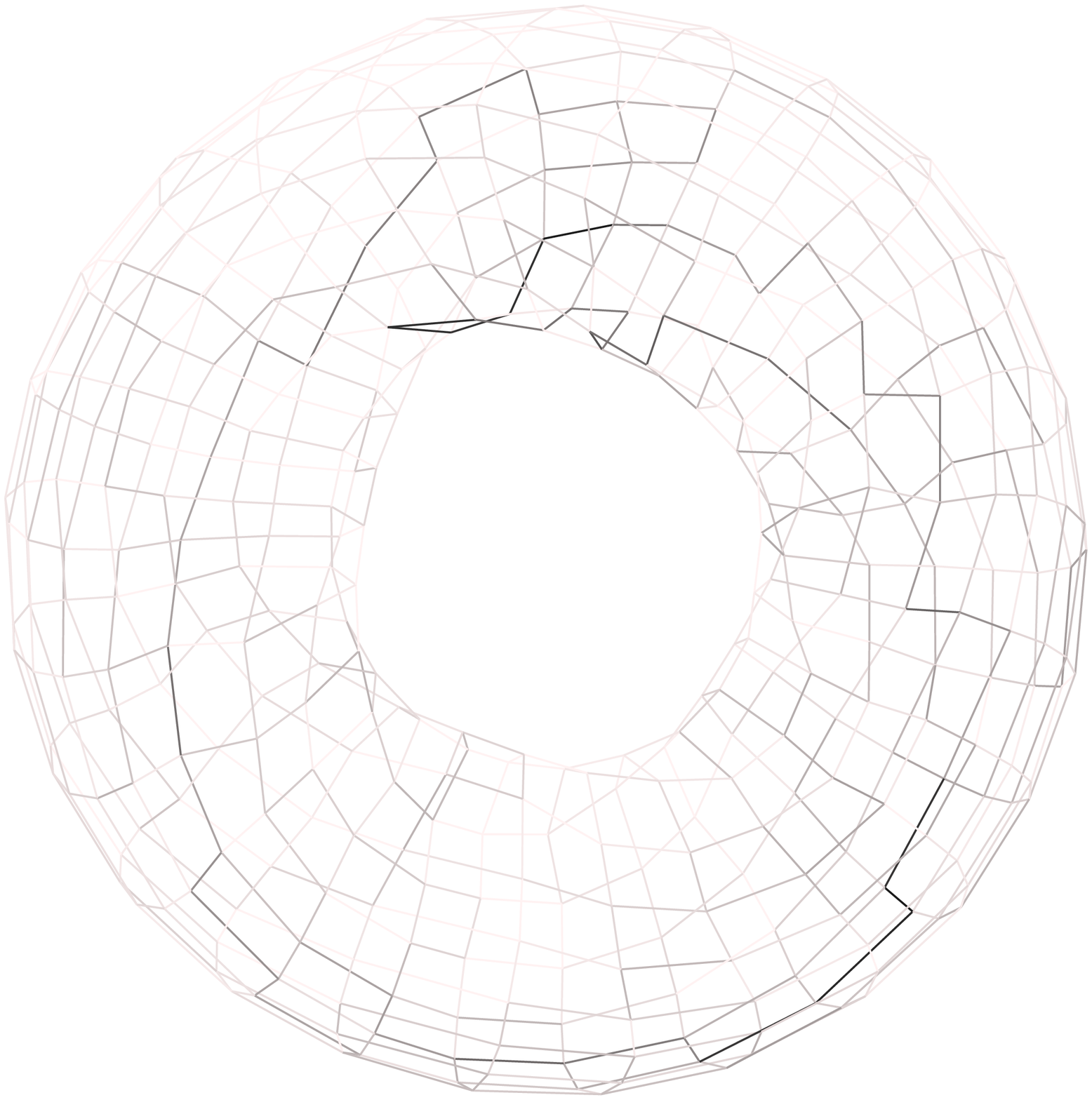}
   \centerline{(a) 120 steps}\medskip
\end{minipage}
\hfill
\begin{minipage}[b]{0.49\linewidth}
  \centering
  \includegraphics[width=\textwidth]{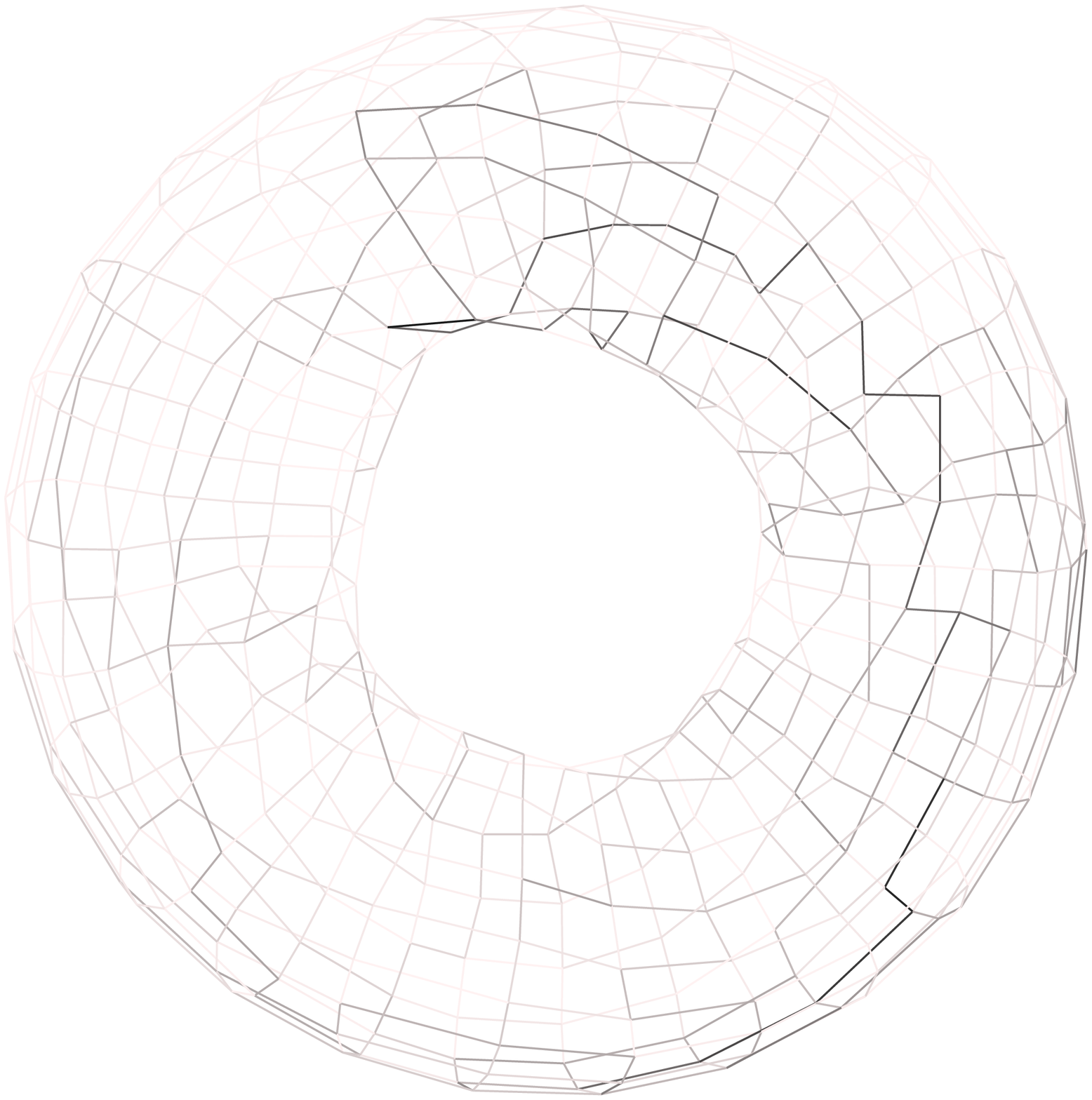}
  \centerline{(b) 180 steps}\medskip
\end{minipage}
\begin{minipage}[b]{0.49\linewidth}
  \centering
 \includegraphics[width=\textwidth]{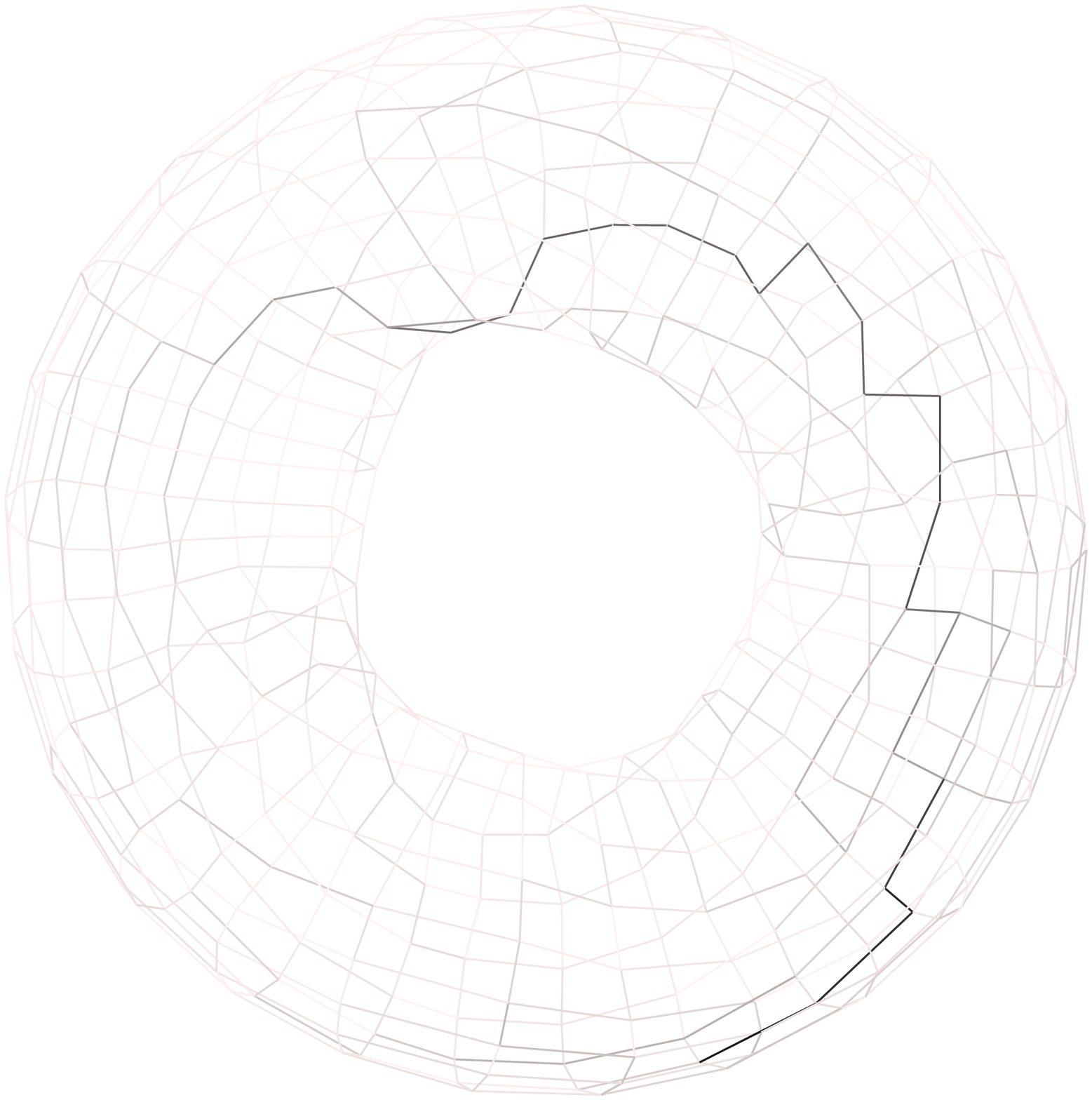}
  \centerline{(c) 230 steps}\medskip
\end{minipage}
\hfill
\begin{minipage}[b]{0.49\linewidth}
\centering
  \includegraphics[width=\textwidth]{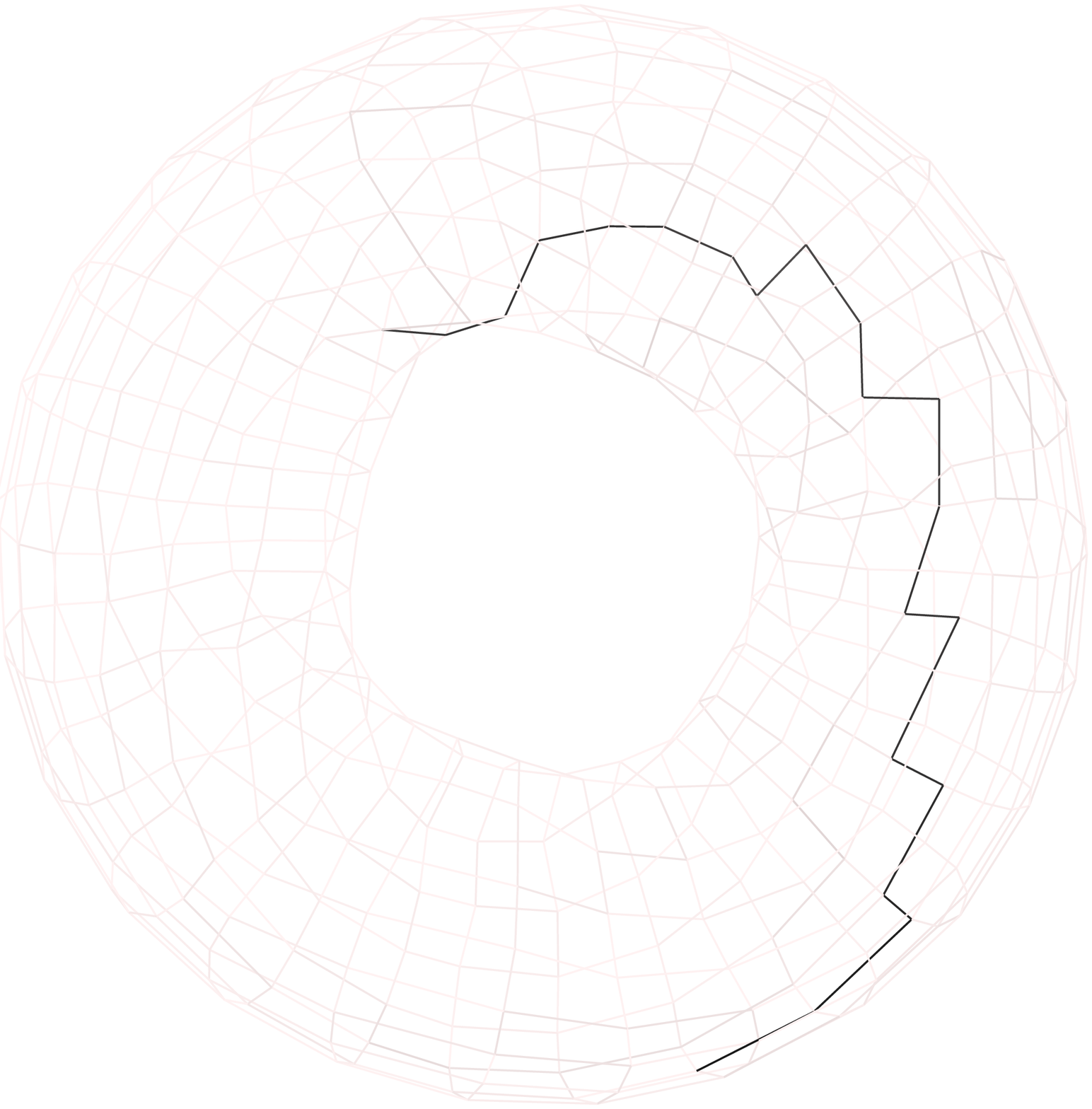}
  \centerline{(d) 360 steps}\medskip
\end{minipage}
\caption{The search of the shortest path between two points of a torus
  graph.  The darker the arcs of the graph, the higher the quantity of
  phe\-ro\-mone. Figure (a) is a picture of the graph at step 120. A
  \emph{step} is a one vertex move for each ant of the colony. Figure
  (b) is a picture of the same graph after 180 steps and so on.}
\label{fig:torusRun}

\end{figure}
\end{center}

Figure \ref{fig:torusRun} shows 4 different states of the graph when
ants are running on it. This figure illustrates the way a path emerges
from the activity of the whole colony. At first ants perform random
walks. When ants discover the destination vertex, they go back to the
nest and lay down the graph a phe\-ro\-mone trail. That why in the
early steps of the run, phe\-ro\-mones are lay approximatively
everywhere on the graph. When ants performing shorter paths they are
faster and can lay down more phe\-ro\-mone trails. Finally a shortest
emerges in the phe\-ro\-mone trails.

This is true that the values of different parameters are highly
related to the structure of the graph


Between two vertices of this graph, the quantity of possible shortest
paths can be important.  In this graph, ants can easily make a lot of
loops without finding the destination vertex. The use of a tabu list
becomes useful in this kind of graph. In effect, the tabu list
prevents ants from making loops in the graph.  Experiments show that
the use of a tabu list that represent the size of the whole path of
the ant gives good results. It is to say that ants are never allowed
to cross already visited nodes.

\begin{figure}[htb]
  \begin{minipage}[b]{1.0\linewidth}
    \centering
    \includegraphics[width=\textwidth]{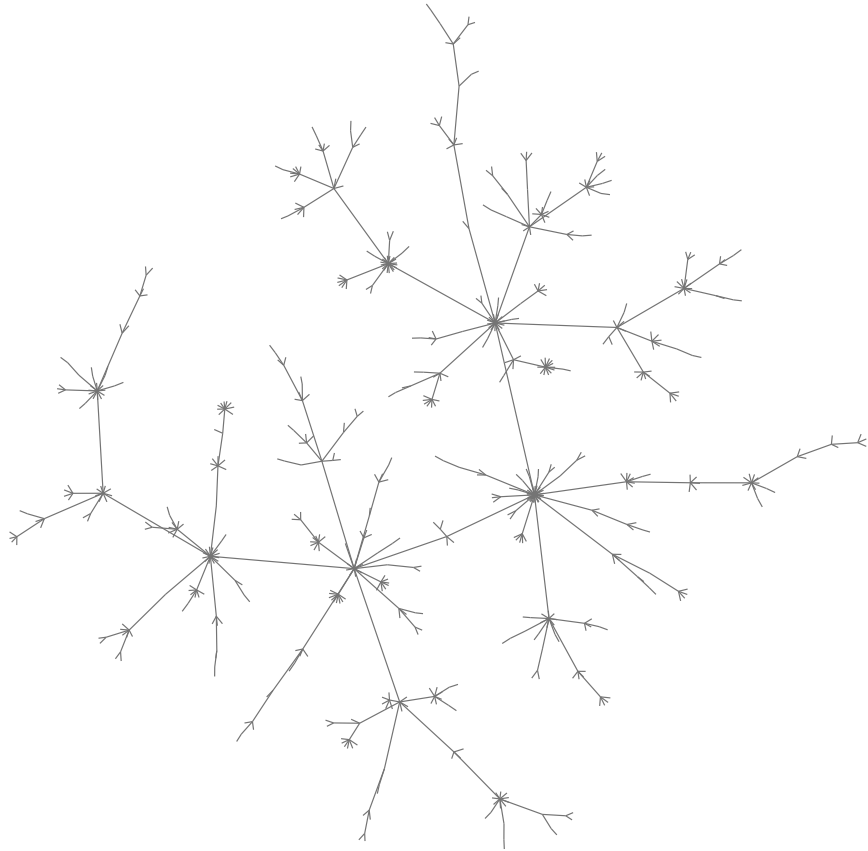}
  \end{minipage}
  \caption{A tree with 750 edges and 751 nodes.}
  \label{fig:tree}
\end{figure}

If it's true that the use of a tabu list is efficient in the case of a
torus graph, but it is not in all circumstances. Let us consider the
graph Fig. \ref{fig:tree}.  In this graph there is no cycle, it is a
tree. If a tabu list is used, ants will loose them self in leafs of
the tree without been able to go back. If no tabu list is use at all,
then strange phenomena occur where some ant go from vertex $A$ to
vertex $B$, then go back to $A$ and so on. Finally the best solution
in this case is to use a tabu list to prevent stagnation and to allow
ants to go back when no more vertices are available. A relatively few
ants are necessary for this graph.

\subsection{Questions Hold by this Approach}

The first question is about the system's ability to produce such
structures.  How can one be sure that the system will produce
structures that have not been defined anywhere in the system? No proof
can be given that such structure will rise. The only evaluation is the
user's point of view when observing the graph.

The method we propose here, as well as the majority of the
metaheuristics, is composed of a set of control parameters that need
to be tuned, more or less according to the problem considered.  Values
taken by these parameters represent a multidimensional space. One
given set of parameter will produce one solution.  If another set of
parameters is given to the system, another solution may appear.  The
question asked deals with the existence of particular conditions
bounded to the values of parameters. Is it possible to make a parallel
between the different sets of parameters and the one dimensional
cellular automata classification made by Stefen Wolfram
\cite{wolfram84universality} where different classes of solutions
exist:
\begin{itemize}
\item Class I: Fixed configurations. All the automaton cells have the
  same state.
\item Class II: Simple structures. Repetitions can be observed.
\item Class III: Chaotic solutions. This kind of automaton usually
  produce fractal structures.
\item Class IV: In this class, complex organizations appear, many
  different structures are observed.
\end{itemize}
Jean-Claude Heudin \cite{heudin98evolution} proposed to compare
Wolfram's classification with experiments made on two dimensional
automata where one parameter can take different values.  In his model,
the different possible values of the $\beta$ parameter produce
different solutions that can be classified with the above scheme. The
$beta$ parameter defines the number of active neighbors around one
particular cell necessary to maintain it in its state.  For one
particular value $\beta=2$, the automaton produced becomes a "game of
life" witch is part of the IV class.  Back to the model presented
here, is it possible to make such a parallel so as to classify the
produced solutions.  In other words, are there particular areas in the
parameter space where the system produces nothing, where it produces
incomprehensible solutions (many different structures, continually
changing) and where it produces understandable solutions (a structure
with the shape of what we are looking for). Finally if such a
comparison can be done, let's call systems with good sets of
parameters, class IV systems.
  
In this case of emerging class IV structures:
\begin{itemize}
\item Are the produced structures stable when a little variation occur
  in the parameter set ? Is the system robust to few changes in the
  parameter space ?
\item What are the characteristics that can be carried out from the
  solutions ?
\item Do all the structures that belong to the class IV share a common
  (or closed) area in the multidimensional space of parameters or are
  there islands of parameters ?
\item In the case of islands, in one particular island, does a little
  change in the parameter lead to a similar solution with little
  differences ?
\end{itemize}

\section{Conclusion}
\label{sec:cs}

 This paper presented an implicit building solution approach using
 emergent properties of ant-based complex systems.  Ant Systems are
 particularly adapted to the observation of emerging structures in an
 environment as solutions to a given problem. This approach is
 relevant especially in case of problems where no global evaluation
 function can be clearly defined and/or when the environment changes.

 The according modeling and solving of two problems illustrated the
 use of this model. It raised the heavy importance that must be given
 to the tuning of some critical parameters. This set of parameters is
 seen as a multidimensional space of possible values.  In this space
 some islands of values may lead to produce wanted structures. If
 these structures are considered as class IV structures, like
 Wolfram's automata, then one part of the future work will consist in
 the identification and the understanding of the areas of parameters.
 
 An analyze of the method was started, it introduced the basis of the
 future work to be done.  The final aim in this project is to
 construct a kind of "grammar" of the different possible structures
 and the different systems running on it.



\bibliographystyle{abbrv} 
\bibliography{Guinand_Pigne}

\end{document}